\documentclass[11pt]{article}

\usepackage[preprint]{acl}

\usepackage{times}
\usepackage{latexsym}

\usepackage[T1]{fontenc}

\usepackage{soul}
\usepackage{xcolor}

\usepackage[utf8]{inputenc}

\usepackage{microtype}

\usepackage{inconsolata}

\usepackage{graphicx}
\usepackage{changepage}

\usepackage{booktabs}       
\usepackage{amsfonts}       
\usepackage{nicefrac}       
\usepackage{microtype}      
\usepackage{graphicx}
\usepackage{amsmath}
\usepackage{amssymb}
\usepackage{array}
\usepackage{multirow}
\usepackage{color}
\usepackage{colortbl}
\usepackage{framed}
\usepackage{bm}
\usepackage{xspace}
\usepackage{enumitem}
\usepackage{xparse}
\usepackage{algorithm}
\usepackage{listings}
\usepackage{url}
\usepackage{mathtools}
\usepackage{pifont}
\usepackage{mathtools}
\usepackage{lipsum}
\usepackage{adjustbox}
\usepackage[x11names]{xcolor}
\usepackage{tcolorbox}
\usepackage[table]{xcolor}
\usepackage{setspace}

\newcommand{\tit}[1]{\smallbreak\noindent\textbf{#1.}}
\newcommand{\tinytit}[1]{\noindent\textbf{#1.}}

\def \ie {\emph{i.e.}}
\def \eg {\emph{e.g.}}

\definecolor{ourcolor}{HTML}{E3DAF0}

\newcommand{\scircled}[1]{\raisebox{-0.15ex}{\scalebox{1.2}{\ding{#1}}}}
\newcommand{\one}[0]{\scircled{172}\xspace}
\newcommand{\two}[0]{\scircled{173}\xspace}
\newcommand{\three}[0]{\scircled{174}\xspace}

\newcommand{\ours}{MixDPO\xspace}
\newcommand{\dataset}{CounterVid\xspace}

\definecolor{promptbg}{gray}{0.95}   
\definecolor{promptborder}{gray}{0.0} 
\definecolor{placeholdercolor}{rgb}{0.1,0.3,0.6}
\definecolor{lightgray}{gray}{0.92}
\newcommand{\placeholder}[1]{\textcolor{placeholdercolor}{\texttt{\{#1\}}}}
\newcommand{\placeholderb}[1]{\textcolor{placeholdercolor}{\texttt{#1}}}

\title{CounterVid: Counterfactual Video Generation for\\Mitigating Action and Temporal Hallucinations in Video-Language Models}

\author{
 \textbf{Tobia Poppi\textsuperscript{1,2,3}},
 \textbf{Burak Uzkent\textsuperscript{1}},
 \textbf{Amanmeet Garg\textsuperscript{1}},
 \textbf{Lucas Porto\textsuperscript{1}},
 \textbf{Garin Kessler\textsuperscript{1}},\\
 \textbf{Yezhou Yang\textsuperscript{1}},
 \textbf{Marcella Cornia\textsuperscript{2}},
 \textbf{Lorenzo Baraldi\textsuperscript{2}},
 \textbf{Rita Cucchiara\textsuperscript{2}},
 \textbf{Florian Schiffers\textsuperscript{1}}
\\
\setstretch{0.5}
 \textsuperscript{1}Amazon Prime Video,
 \textsuperscript{2}University of Modena and Reggio Emilia,
 \textsuperscript{3}University of Pisa
\\
\setstretch{0.5}
\texttt{\{tobipop,burauzke,amanmega,lporto,kesslerg,imyzyang,floschi\}@amazon.com,}\\
\texttt{\{name.surname\}@unimore.it, \{name.surname\}@phd.unipi.it}
}

\begin{document}
\maketitle

\begin{abstract}
Video-language models (VLMs) achieve strong multimodal understanding but remain prone to hallucinations, especially when reasoning about actions and temporal order. Existing mitigation strategies, such as textual filtering or random video perturbations, often fail to address the root cause: over-reliance on language priors rather than fine-grained visual dynamics. We propose a scalable framework for \emph{counterfactual video generation} that synthesizes videos differing only in actions or temporal structure while preserving scene context. Our pipeline combines multimodal LLMs for action proposal and editing guidance with diffusion-based image and video models to generate semantic hard negatives at scale. Using this framework, we build \dataset, a synthetic dataset of $\sim$26k preference pairs constructed from short counterfactual action clips and targeting both action recognition and controlled action-sequence ordering. We further introduce \ours, a unified Direct Preference Optimization approach that jointly leverages textual and visual preferences. Across Qwen2.5-VL and InternVL3 backbones, \ours substantially improves action recognition and temporal ordering, yields gains on most standard video hallucination benchmarks, and largely preserves general video understanding. Our source code, trained models, and dataset are available at {\small\url{https://aimagelab.github.io/CounterVid}}.
\end{abstract}

\section{Introduction}
\label{sec:intro}

Large-scale video-language models (VLMs)~\cite{liu2023visual,liu2024improved,chen2024internvl,zhu2023minigpt} have demonstrated strong performance on open-ended video understanding tasks~\cite{maaz2024video,Ding_2024_CVPR,li2023videochat,lin2023videollava,liu2024llavanext,zhang2024llavanextvideo}. Despite this progress, these models remain susceptible to hallucinations, producing responses that are not supported by visual evidence~\cite{liu2024survey,wang2024videohallucer,bai2024hallucination}.

Video hallucinations manifest in diverse forms, ranging from static appearance errors, such as describing nonexistent objects or misidentifying attributes, to dynamic failures caused by incorrect temporal understanding. Recent studies~\cite{wang2024videohallucer,im2025vidhalluc} show that modern VLMs often confuse visually similar actions or infer event sequences from language priors rather than observed motion. For example, a model may correctly recognize objects and scenes yet incorrectly assert that an action occurred, or assume a typical order of events even when contradicted by the video. Notably, such errors persist even in strong instruction-tuned models, highlighting limitations beyond surface-level reasoning.

\begin{figure}[t]
    \centering
    \includegraphics[width=0.99\linewidth]{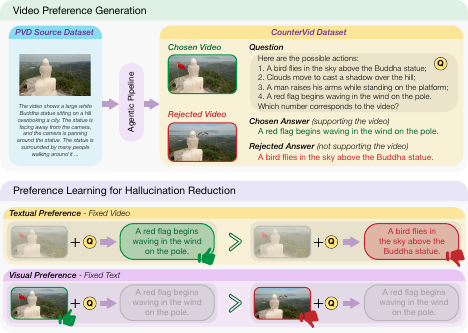}
    \vspace{-0.6cm}
    \caption{\textbf{Video preference generation and learning.} From a source video-caption pair, our pipeline creates semantically matched chosen/rejected videos and answers for textual and visual preference learning.}
    \label{fig:teaser}
    \vspace{-0.3cm}
\end{figure}

Motivated by these observations, we focus on two critical and under-explored failure modes: \textit{semantic action misidentification} and \textit{incorrect inference of event order}.
Existing approaches address video hallucination by introducing additional supervision, either through architectural modifications~\cite{ma2024vista_llama,eventhallusion2024} or preference-based learning~\cite{rafailov2023dpo,ding2025pami_vdpo}. However, many existing methods rely on negative samples from random perturbations, such as frame shuffling. While such perturbations disrupt temporal coherence, they rarely produce \emph{semantic} counterfactuals, \ie, videos that preserve scene context while differing meaningfully in actions or event order. Consequently, models may succeed by exploiting low-level artifacts or static cues, rather than learning fine-grained action dynamics~\cite{huang2018makes}.

To overcome these limitations, we introduce a scalable framework for \emph{counterfactual video generation} that produces short action clips targeting action recognition and controlled action-sequence ordering. Instead of random augmentations, our approach constructs videos that differ only in their underlying action trajectories while preserving scene identity. The proposed pipeline integrates complementary generative components: multimodal LLMs~\cite{minaee2024large,caffagni2024revolution} for plausible alternative actions and structured edits, image editing models for action-completion frames~\cite{wu2025qwenimage,labs2025flux}, and image-to-video diffusion models~\cite{wan2025wan} for generating temporally coherent clips. This modular design disentangles semantic reasoning from visual synthesis, while enabling precise control over the generated counterfactuals.

Using this framework, we compile \textbf{\dataset}, a large-scale dataset of approximately 26k preference pairs built from short synthetic clips with controlled action variations. Fig.~\ref{fig:teaser} summarizes the preference-data construction and learning setup. From each anchor scene, we generate multiple action-consistent clips and construct preference pairs targeting both action recognition and temporal ordering, spanning free-form, binary, and structured formats. These data support two complementary supervision signals: \emph{textual preferences}, contrasting grounded and hallucinated answers under fixed visual input, and \emph{visual preferences}, contrasting correct and counterfactual videos under fixed textual query and answer.

We leverage these signals through \textbf{\ours}, a simple framework based on Direct Preference Optimization (DPO)~\cite{rafailov2023dpo} that integrates textual and visual supervision. Its contribution is to formulate visual preferences over same-scene semantic counterfactuals and optimize them jointly with standard answer-side preferences. By jointly promoting
output grounding and sensitivity to visual evidence, our approach directly targets the mechanisms underlying video hallucination. Experiments on Qwen2.5-VL~\cite{bai2025qwen2} and InternVL3~\cite{zhu2025internvl3} backbones show that fine-tuning with \ours yields substantial improvements in action recognition and temporal ordering, with positive transfer to established benchmarks such as EventHallusion~\cite{eventhallusion2024} and VidHalluc~\cite{im2025vidhalluc}.

\noindent\textbf{Contributions.} Our contributions are threefold:
\begin{itemize}[leftmargin=*, noitemsep, topsep=0pt]
    \item We introduce a modular counterfactual video generation pipeline that combines language-guided action proposal with diffusion-based generation to produce semantic hard negatives at scale.
    \item We construct \dataset, a synthetic preference dataset designed to expose action-recognition and temporal-ordering hallucinations under controlled visual context.
    \item We present \ours, a unified preference-based alignment framework that jointly leverages textual and visual supervision to improve grounding and temporal sensitivity in VLMs.
\end{itemize}

\section{Related Work}
\tinytit{Video Hallucinations}
Recent works have shown that VLMs frequently generate hallucinated outputs that are not grounded in visual evidence, despite strong performance on standard video understanding benchmarks~\cite{bai2024hallucination,liu2024survey,wang2024videohallucer}.
VideoHallucer~\cite{wang2024videohallucer} provides an initial taxonomy, distinguishing intrinsic hallucinations that misdescribe content present in the video from extrinsic ones that fabricate absent content.
Subsequent studies emphasize temporal understanding as a particularly vulnerable dimension: models often confuse visually similar actions or infer event order from language priors rather than observed motion~\cite{im2025vidhalluc,eventhallusion2024}.
Together, these works indicate that action recognition and temporal reasoning remain central challenges for VLMs.

\tit{Architectural and Decoding Strategies}
Several approaches mitigate hallucinations by strengthening visual grounding through architectural or inference-time interventions.
VISTA-LLAMA~\cite{ma2024vista_llama} enforces balanced attention over visual and textual tokens to reduce the tendency to ignore video input.
Decoding-based methods~\cite{eventhallusion2024,leng2024vcd}, instead, counteract language priors by contrasting outputs generated from original and degraded or manipulated visual inputs. Self-refinement strategies, including Volcano~\cite{lee2024volcano}, further encourage models to verify and revise predictions against visual evidence~\cite{wang2025grounded,guo2025trace}.
While effective, these methods typically require custom architectures, additional inference passes, or specialized decoding schemes, which can limit scalability.

\begin{figure*}[t]
    \centering
    \includegraphics[width=0.99\linewidth]{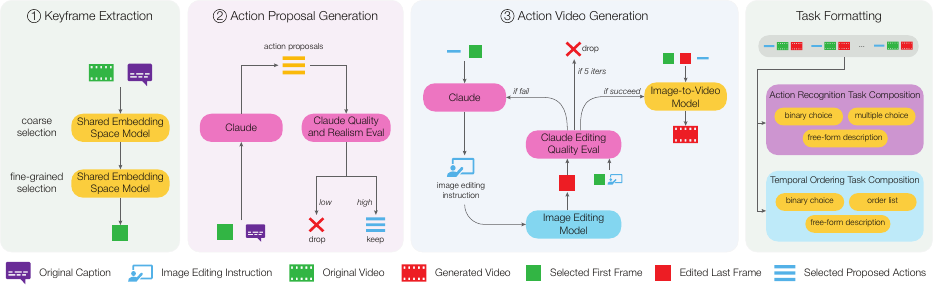}
    \vspace{-0.15cm}
    \caption{\textbf{Overview of the counterfactual generation framework.} Starting from a real video-caption pair, we extract a representative keyframe with a shared embedding space model, propose multiple alternative actions with a multimodal LLM, synthesize one counterfactual video per proposed action via image editing and image-to-video generation, and finally compose preference pairs for action recognition and temporal ordering.}
    \label{fig:pipeline}
    \vspace{-0.3cm}
\end{figure*}

\tit{Preference-Based Alignment}
A different line of work reduces hallucinations through preference-based alignment. In the image domain, V-DPO~\cite{xie2024vdpo} and mDPO~\cite{wang2024mdpo} extend DPO~\cite{rafailov2023dpo} with both output-level preferences (same input, different answers) and input-level preferences (different inputs, same query).
In the video domain, PaMi-VDPO~\cite{ding2025pami_vdpo} constructs preference pairs using video augmentations such as frame shuffling, cropping, and temporal reversal, while related studies apply preference optimization on synthetic videos to reduce commonsense and physics-related hallucinations~\cite{li2025videohallu}.

Despite their effectiveness, augmentation-based approaches are limited by the source content: random perturbations can disrupt temporal coherence but rarely produce counterfactuals that preserve scene context while varying actions or event order. As noted in~\cite{ding2025pami_vdpo}, the most informative negatives are visually similar yet semantically distinct, which simple augmentations cannot reliably generate. Our work addresses this gap by synthesizing content-aware counterfactual videos with controlled action and temporal variations, and by jointly leveraging visual and textual preferences to promote visual grounding over language priors.

\section{Proposed Method}
\label{sec:method}

We propose a scalable framework that (i) synthesizes \emph{counterfactual} videos that differ only in action dynamics while preserving scene context, and (ii) aligns a VLM using a unified preference objective that couples \emph{textual} and \emph{visual} supervision.

\subsection{Task Definition and Preference Data}
\label{sec:taskdef}

\tinytit{Video QA with Hallucinations}
Let $V$ denote a video clip, $Q$ a question, and $A$ a free-form textual answer. A VLM with parameters $\theta$ defines a conditional distribution $\pi_\theta(A\,|\,V,Q)$.
A \emph{hallucination} occurs when $A$ is not supported by the visual evidence in $V$, despite being linguistically plausible. Using short action clips and controlled concatenations of them, we focus on two dynamic failure modes:
(i) \textbf{action recognition} (misidentifying the action in an otherwise correct response) and
(ii) \textbf{temporal ordering} (incorrectly inferring the order of concatenated actions). This controlled setting isolates order sensitivity.

\tit{Preference Supervision}
Our training signal uses pairwise preferences of the form $(x, y^+, y^-)$ or ($x^+, x^-, y)$, where $x$ is the multimodal context, $y$ is the response, and $+$ and $-$ superscripts indicate preferred and rejected alternatives. We construct two complementary preference types.

\tit{Textual Preferences (t-pref)}
For a fixed context $x=(V,Q)$, we prefer a grounded answer $A^+$ over a hallucinated answer $A^-$:
\begin{equation}
    (x, y^+, y^-) = \big((V,Q), A^+, A^-\big).
    \label{eq:tpref_def}
\end{equation}

\tit{Visual Preferences (v-pref)}
For a fixed text pair $(Q,A)$, we prefer the \emph{correct} visual context over a \emph{counterfactual} one:
\begin{equation}
    (x^+, x^-, y) = \big((V^+,Q), (V^-,Q), A\big),
    \label{eq:vpref_def}
\end{equation}
where $V^+$ and $V^-$ depict different actions (or different action sequences) under nearly identical scene context. This directly penalizes ``blind'' generation, where $\pi_\theta(A\,|\,V^+,Q)\approx\pi_\theta(A\,|\,V^-,Q)$.

\subsection{Counterfactual Video Generation Pipeline}
\label{sec:pipeline}
We start from a large collection of real video–caption pairs $(V^{\text{real}}, C^{\text{real}})$ and use them only to extract an anchor frame; all preference videos are \emph{synthetic}.
Given one anchor frame, we synthesize a set of $N$ action videos
$\mathcal{G}=\{(V_i, C_i)\}_{i=1}^N$
that share the same static context (scene, objects, viewpoint) but differ in the action dynamics.
These generated videos form the basis for both action-recognition and temporal-ordering preference pairs. An overview of the generation pipeline is shown in Fig.~\ref{fig:pipeline}, and we elaborate on each part in the following sections. 

\tit{\one Keyframe Extraction}
For each real pair $(V^{\text{real}}, C^{\text{real}})$, we select a representative keyframe $I^\text{start}$ that best matches the caption and provides a stable starting state for editing and generation.
We use a coarse-to-fine retrieval approach using a shared embedding space model~\cite{zhai2023sigmoid}:
(i) \emph{coarse} sampling at 2 fps to identify a high-alignment temporal neighborhood, and
(ii) \emph{refined} sampling at 12 fps within that neighborhood to choose the final keyframe.
We compute caption-frame similarity by averaging over multiple spatial crops (center/left/right) to reduce sensitivity to framing and to better capture the global scene.

\tit{\two Action Proposal and Filtering}
Conditioned on the keyframe $I^\text{start}$ and the original caption $C^{\text{real}}$, we prompt a multimodal LLM (\ie, Claude~\cite{claude4}) to propose $N$ alternative actions
$\{C_i\}_{i=1}^N$ that are:
(i) \emph{plausible} given the scene state in $I^\text{start}$,
(ii) \emph{distinct} from each other (to ensure hard negatives), and
(iii) \emph{visually expressible} within a short clip.
We then filter proposals using the same LLM with explicit criteria for realism, feasibility, and action clarity (templates are in Appendix~\ref{sec:prompts}).

\tit{\three Action Video Generation}
For each action caption $C_i$, we generate an \emph{end frame} $I_i^{\text{end}}$ depicting the completion of the action starting from $I^\text{start}$.
We first convert $C_i$ into structured editing instructions, then apply an image editing model~\cite{wu2025qwenimage} to obtain $I_i^{\text{end}}$.
Because image editing can introduce semantic drift and artifacts, we employ an iterative refinement loop:
after each attempt, the multimodal LLM evaluates whether $I_i^{\text{end}}$ matches the intended action and preserves the scene identity; if not, it revises the editing prompt.

Finally, we synthesize a temporally coherent video $V_i$ with an image-to-video model~\cite{wan2025wan} conditioned on $(I^\text{start}, I_i^{\text{end}}, C_i)$.
This yields a set of action-consistent clips $\{V_i\}_{i=1}^N$ that share the same context by construction (same anchor frame) and differ in the action trajectory.
We further apply a post-generation safety screening step using visual safety classifiers. Samples flagged as potentially unsafe are removed from the dataset. More details are reported in Appendix~\ref{sec:training-details-addl}.

\subsection{Task and Preference Construction}
After generation, we discard $(V^{\text{real}}, C^{\text{real}})$, and construct samples for \dataset\ exclusively from the generated set
$\mathcal{G}=\{(V_i, C_i)\}_{i=1}^N$ obtained from the same anchor frame. Preference pairs are generated for different target capabilities (\ie, action recognition and temporal ordering) and in different formats (free-form, binary, and multiple choice or order list depending on the sample type). In the following, let $Q_{\text{act}}$ denote a generic action query (\eg, ``What action is shown?'') and let $\mathcal{C}=\{C_i\}_{i=1}^N$ be the set of action captions proposed for the anchor.

\subsubsection{Preference Pairs for Action Recognition}
In action recognition, we employ a single generated clip $V_i$ and ask the model to identify its action. To generate t-pref pairs, we keep the input context fixed, $x=(V_i, Q_{\text{act}})$, and contrast a grounded caption with a plausible but incorrect one sampled from the same anchor set:
\begin{equation}
    A^+=C_i,\;A^-=C_j,\;j\sim\text{Unif}(\{1..N\}\setminus i).
    \label{eq:actrec_tpref}
\end{equation}

For v-pref pairs, we fix the text pair $(Q_{\text{act}}, A=C_i)$ and swap the video, so that the same answer is correct for one clip and false for the other:
\begin{equation}
    (V^+,V^-)=(V_i,V_j),\;j\sim\text{Unif}(\{1..N\}\setminus i).
    \label{eq:actrec_vpref}
\end{equation}

\tinytit{Formats}
Free-form uses $Q_{\text{act}}$ with $A$ as caption. Binary choice presents one candidate and asks whether it matches the video. Multiple choice presents the candidate set $\mathcal{C}$, and the answer is the correct option index. In all cases, the preferred/rejected elements are defined via Eq.~\eqref{eq:actrec_tpref} or Eq.~\eqref{eq:actrec_vpref}.

\subsubsection{Preference Pairs for Temporal Ordering}
Temporal ordering is evaluated under minimal contextual change. Both chosen and rejected sequences are formed by concatenating generated clips from the same anchor frame.

\tit{Sequence Construction}
We sample $K$ distinct actions $\{i_1,\dots,i_K\}\subset\{1..N\}$ and concatenate the corresponding clips $S = [V_{i_1};V_{i_2};\dots;V_{i_K}]$, where $[\cdot;\cdot]$ denotes temporal concatenation. We define a \emph{chosen} order $S^+$ as the sequence used to construct the concatenated video, and a \emph{rejected} order $S^-$ by applying a non-identity random permutation.

\begin{figure*}[t]
    \centering
    \includegraphics[width=0.99\linewidth]{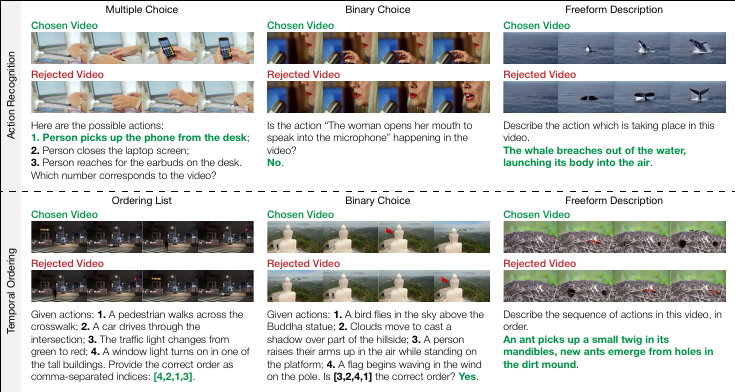}
    \vspace{-0.15cm}
    \caption{\textbf{Qualitative examples from \dataset}, illustrating action recognition and temporal ordering samples across multiple-choice, order-list, binary, and free-form formats. All videos and answers are generated.}
    \label{fig:ex-dataset}
    \vspace{-0.3cm}
\end{figure*}

\tit{Preference Pairs Generation}
For t-pref pairs, we keep the context fixed, $x=(S^+, Q_{\text{seq}})$, and contrast a grounded description with a hallucinated one obtained by caption swaps within the same anchor set.
Concretely, $A^+$ describes the action order in $S^+$, while $A^-$ is formed by permuting or substituting one or more action captions from the same anchor set.
For v-pref pairs, we fix the textual target to the correct order and swap the visual evidence between $S^+$ and $S^-$. This directly enforces that the model assigns a higher likelihood to the correct ordering when the underlying clips are permuted but visually similar in context.

\tit{Formats} Free-form uses $Q_{\text{seq}}$ with $A$ as an ordered list of actions. Binary choice presents a candidate ordering and asks whether it matches the video sequence. Order list presents the unordered action set and requires the model to output the correct temporal order as a sequence of indices.

\smallskip
\noindent Representative generated preference pairs from \dataset are shown in Fig.~\ref{fig:ex-dataset}.

\subsection{The \ours Alignment Framework}
We fine-tune a base VLM with a unified preference objective that combines textual and visual signals.
Our goal is not only to improve answer quality, but to increase \emph{input sensitivity}: answers should change when the visual evidence changes.

\tit{Preliminaries}
DPO~\cite{rafailov2023dpo} optimizes a policy (\ie, the video-language model being fine-tuned) $\pi_\theta$ given preference pairs $(x, y^+, y^-)$ without an explicit reward model. Let $\pi_{\text{ref}}$ be a frozen reference model, DPO defines an implicit reward as the log-likelihood ratio between the trainable policy and the reference model: $r_\theta(x, y) = \log \frac{\pi_\theta(y|x)}{\pi_{\text{ref}}(y|x)}$. The overall DPO objective is then defined as:
\begin{equation}
\mathcal{L}_{\text{DPO}} = -\mathbb{E}\left[\log \sigma\left(\beta \left(r_\theta(x, y^+) - r_\theta(x, y^-)\right)\right)\right],
\label{eq:dpo}
\end{equation}
where $\sigma$ is the sigmoid, and $\beta$ controls the strength of the deviation from $\pi_{\text{ref}}$.

\tit{Unified MixDPO for Hallucination Mitigation} As multimodal hallucination often stems from the model ignoring the visual context, we define a unified loss that integrates two complementary signals built on the two aforementioned preference types:
\begin{equation}
\mathcal{L}_{\text{MixDPO}}(\theta) = \mathcal{L}_{\text{t-pref}}(\theta) + \lambda \mathcal{L}_{\text{v-pref}}(\theta).
\label{eq:mixdpo}
\end{equation}
Here, we use $\lambda=1$ as a balanced default between answer discrimination and visual grounding. Section~\ref{sec:robustness} evaluates the sensitivity to this choice and to the textual/visual preference composition.

\tit{Textual Preference Loss} 
We force the model to sharpen its probability distribution over the correct tokens while suppressing plausible but visually unsupported sequences. Formally, given a textual preference triple $\big((V,Q),A^+,A^-\big)$, we define
\begin{equation}
\begin{split}
\mathcal{L}_{\text{t-pref}}=
-\mathbb{E}\Big[\log \sigma\big(\beta (r_\theta((V,Q),A^+)- \\
r_\theta((V,Q),A^-))\big)\Big].
\end{split}
\label{eq:tpref}
\end{equation}
In \dataset, $A^-$ is constructed as an \emph{action-caption swap} from the same anchor set, yielding a fluent but visually unsupported alternative, directly targeting language-prior guessing.

\tit{Visual Preference Loss} 
For a visual preference tuple $\big((V^+,Q),A,(V^-,Q),A\big)$, we define
\begin{equation}
\begin{split}
\mathcal{L}_{\text{v-pref}}=
-\mathbb{E}\Big[\log \sigma\big(\beta (r_\theta((V^+,Q),A) \\ -r_\theta((V^-,Q),A))\big)\Big].
\end{split}
\label{eq:vpref}
\end{equation}
This term explicitly enforces that the same answer $A$ should be assigned a high likelihood only under the correct visual evidence.
When a model ignores video input, the two likelihoods become similar; maximizing their margin forces the policy to attend to discriminative temporal cues.

Minimizing $\mathcal{L}_{\text{MixDPO}}$ teaches the model both
(i) \emph{output grounding} (prefer grounded over hallucinated answers under fixed video) and
(ii) \emph{input sensitivity} (prefer the correct video over a counterfactual under fixed text),
which is essential for mitigating temporal hallucinations in VLMs.

\section{Experiments}
\subsection{Experimental Setup}

\tinytit{\dataset Details}
The final version of \dataset consists of 26,167 preference pairs spanning action recognition and temporal ordering, used to optimize VLMs with \ours. Action recognition includes 12,919 samples, distributed across free-form (4,416), binary-choice (4,416), and multiple-choice (4,087) formats. Temporal ordering consists of 13,248 samples, evenly split among free-form, order-list, and binary-choice queries (4,416 each). We maintain a consistent preference ratio of 70\% visual preferences (v-pref) and 30\% textual preferences (t-pref). In addition to training data, we curate 2,910 held-out samples for benchmarking VLMs under the same evaluation settings. We perform the train-test split at the anchor level, ensuring preference pairs derived from the same anchor frame do not appear in both sets.

\tit{Data Generation Settings}
We generate counterfactual videos using the pipeline described in Sec.~\ref{sec:pipeline}. Real anchor frames and captions are selected from the PE Video Dataset (PVD)~\cite{bolya2025perception}, from which we only select clips shorter than 10 seconds, which are likely to contain a single action. Keyframe extraction is performed with SigLIP-SO400M~\cite{zhai2023sigmoid}. Action proposal and filtering use Claude-4-Sonnet~\cite{claude4}.
We use Claude as one implementation of the multimodal planner/evaluator in our modular pipeline. Appendix~\ref{sec:supp_results} evaluates whether its filtering decisions are consistent with open-source VLM judges.
End-frame generation is carried out via Qwen-Image-Edit~\cite{wu2025qwenimage} with an iterative refinement loop (up to five attempts) to ensure action correctness and scene consistency. Final videos are synthesized using Wan2.2-I2V-14B~\cite{wan2025wan}, producing clips of approximately three seconds at 680$\times$384 resolution. Additional details are provided in the Appendix.

\tit{Baselines}
We evaluate \ours across two VLM families: Qwen2.5-VL-3B and Qwen2.5-VL-7B~\cite{bai2025qwen2}, and InternVL3-9B~\cite{zhu2025internvl3}. For each backbone, \textbf{Base} denotes the unfine-tuned model, \textbf{T-pref-DPO} applies DPO~\cite{rafailov2023dpo} using only textual preference pairs to isolate answer-level supervision, and \textbf{SFT} fine-tunes on original PVD captions~\cite{bolya2025perception} to control for domain exposure. We also compare with \textbf{PaMi-VDPO}~\cite{ding2025pami_vdpo}, a video-preference baseline whose chosen video and Crop/Shuffle negatives are jointly processed through its prompt-aware multi-instance objective. We use a matched three-epoch budget, and the main-table results use \dataset videos as the chosen examples. To examine whether the observed behavior extends to a larger model, Appendix~\ref{sec:scale_14b} reports a matched \dataset-only experiment on InternVL3-14B. Appendix~\ref{sec:training-details-addl} provides full implementation details, while Appendix~\ref{sec:supp_results} compares variants using PVD versus \dataset chosen videos.

\tit{Evaluation Benchmarks} 
We evaluate our method on the held-out \dataset split, a targeted in-family benchmark for action recognition and controlled short action-sequence ordering, and on a set of established video hallucination benchmarks.
Specifically, we conduct experiments on VideoHallucer~\cite{wang2024videohallucer}, VidHalluc~\cite{im2025vidhalluc}, EventHallusion~\cite{eventhallusion2024}, and VideoHallu~\cite{li2025videohallu}, which cover hallucinations related to object existence, action recognition, temporal ordering, and event-level reasoning. To verify that hallucination mitigation does not degrade general video understanding, we additionally evaluate on VideoMME~\cite{fu2024videomme}, NExT-QA~\cite{xiao2021nextqa}, and TempCompass~\cite{liu2024tempcompass}. Further details on evaluation benchmarks and settings are provided in Appendix~\ref{sec:benchmark-details}.

\begin{table}[t]
\centering
\setlength{\tabcolsep}{0.3em}
\resizebox{\linewidth}{!}{
\begin{tabular}{lccccc}
\toprule
\textbf{Axis / Label} & \textbf{Free-Form} & \textbf{Binary} & \textbf{Multi-Choice} & \textbf{Order} & \textbf{All} \\
\midrule
\multicolumn{6}{l}{\textit{Label validity}} \\
\hspace{0.4cm}Wrong      & 5.0\%  & 6.9\% & 2.7\%  & 7.0\%  & 5.4\% \\
\hspace{0.4cm}Ambiguous     & 8.1\%  & 4.9\% & 2.3\%  & 11.9\% & 6.8\% \\
\hspace{0.4cm}Correct    & 86.9\% & 88.2\% & 95.0\% & 81.1\% & 87.8\% \\
\midrule
\multicolumn{6}{l}{\textit{Visual quality}} \\
\hspace{0.4cm}Very Poor  & 2.0\%  & 2.3\% & 2.4\%  & 2.5\%  & 2.3\% \\
\hspace{0.4cm}Poor       & 10.5\% & 9.5\% & 6.4\%  & 12.4\% & 9.7\% \\
\hspace{0.4cm}Acceptable    & 21.4\% & 21.6\% & 21.7\% & 20.9\% & 21.4\% \\
\hspace{0.4cm}Good       & 58.1\% & 59.1\% & 62.2\% & 56.2\% & 58.9\% \\
\hspace{0.4cm}Excellent  & 8.0\%  & 7.5\% & 7.3\%  & 8.0\%  & 7.7\% \\
\midrule
\# Samples & 108 & 103 & 46 & 43 & 300 \\
\bottomrule
\end{tabular}
}
\vspace{-0.1cm}
\caption{\textbf{Human evaluation by task format.} We assess label validity and visual quality separately to avoid conflating semantics with artifacts.}
\label{tab:human_eval_tasks}
\vspace{-0.2cm}
\end{table}

\subsection{Data Quality and Pipeline Evaluation}
\label{subsec:human_eval_new}

To assess the quality of the automatically generated \dataset, we conduct a human evaluation on a representative subset of the held-out split. We evaluate two complementary axes: (i) preference-label validity (\ie, whether the chosen video provides stronger visual support for the answer than the rejected one), and (ii) visual quality of the generated counterfactual videos.

\tit{Evaluation Protocol}
We randomly sample 300 examples from the held-out set ($\approx$10.3\% of 2{,}910 samples), with proportional coverage across free-form (36.0\%), binary-choice (34.3\%), multiple-choice (15.3\%), and order-list (14.3\%) formats. Each sample is independently reviewed by ten evaluators, who inspect the generated videos together with the associated questions and answers. To evaluate video correctness, evaluators choose among correct, ambiguous, and wrong; for visual quality, they rate the video as excellent, good, acceptable, poor, or very poor.
Additional annotation protocol details are given in Appendix~\ref{sec:training-details-addl}.

\tit{Results and Analysis}
Table~\ref{tab:human_eval_tasks} reports the resulting breakdown. Overall, 87.8\% of samples are judged correct and only 5.4\% wrong, indicating that preference-label errors are uncommon. Under the quality axis, 66.6\% of videos are rated good or excellent, and 88.0\% are at least acceptable. Only 12.0\% are poor or very poor, with severe degradation accounting for 2.3\%. These results suggest that most generated counterfactuals are sufficiently interpretable for semantic preference learning, while the two-axis annotation makes label noise and visual artifacts explicit. Appendix~\ref{sec:failure_analysis} analyzes the 46 unique samples flagged for an incorrect or ambiguous label, poor visual quality, or both, and Fig.~\ref{fig:generation_success_failures} shows representative successes and failures.

\begin{table}[t]
  \centering
  \setlength{\tabcolsep}{0.35em}
  \resizebox{\linewidth}{!}{
  \begin{tabular}{@{}p{0.6cm}@{}l c ccc c ccc c}
    \toprule
    \multicolumn{2}{l}{} && \multicolumn{3}{c}{\textbf{Temporal Ord.}} && \multicolumn{3}{c}{\textbf{Action Rec.}} & \\
    \cmidrule{4-6} \cmidrule{8-10}
    \multicolumn{2}{l}{\textbf{Model}} && FF & OL & BC && FF & MC & BC & \textbf{Avg} \\
    \midrule
    \rowcolor{lightgray}
    \multicolumn{2}{l}{$\blacktriangledown$ \textit{Qwen2.5-VL-3B}} && 29.8 & 1.6 & 48.9 && 28.0 & 48.4 & 64.6 & 36.9 \\
    & SFT && 29.4 & 2.0 & 47.9 && 27.2 & 51.9 & 65.2 & 37.2 \\
    & T-pref-DPO && 32.2 & 15.7 & 56.0 && \textbf{29.6} & 58.7 & 69.5 & 43.6 \\
    & PaMi-VDPO && \textbf{32.9} & 1.4 & 47.5 && 29.1 & 52.3 & 68.0 & 38.5 \\
    & \cellcolor{ourcolor}\textbf{\ours (Ours)}
      & \cellcolor{ourcolor}
      & \cellcolor{ourcolor}31.0
      & \cellcolor{ourcolor}\textbf{16.7}
      & \cellcolor{ourcolor}\textbf{64.0}
      & \cellcolor{ourcolor}
      & \cellcolor{ourcolor}29.5
      & \cellcolor{ourcolor}\textbf{60.0}
      & \cellcolor{ourcolor}\textbf{70.3}
      & \cellcolor{ourcolor}\textbf{45.2} \\
    \midrule
    \rowcolor{lightgray}
    \multicolumn{2}{l}{$\blacktriangledown$ \textit{Qwen2.5-VL-7B}} && 70.3 & 16.5 & 57.8 && 55.1 & 68.6 & 78.2 & 57.8 \\
    & SFT && 69.3 & 14.9 & 55.9 && 55.3 & 67.7 & 79.2 & 57.1 \\
    & T-pref-DPO && \textbf{72.2} & 37.1 & 67.4 && 56.0 & 70.8 & 79.8 & 63.9 \\
    & PaMi-VDPO && 42.2 & 13.4 & 57.6 && 32.9 & 64.4 & 74.5 & 47.5 \\
    & \cellcolor{ourcolor}\textbf{\ours (Ours)}
      & \cellcolor{ourcolor}
      & \cellcolor{ourcolor}\textbf{72.2}
      & \cellcolor{ourcolor}\textbf{43.8}
      & \cellcolor{ourcolor}\textbf{72.7}
      & \cellcolor{ourcolor}
      & \cellcolor{ourcolor}\textbf{56.3}
      & \cellcolor{ourcolor}\textbf{71.2}
      & \cellcolor{ourcolor}\textbf{80.7}
      & \cellcolor{ourcolor}\textbf{66.2} \\
    \midrule
    \rowcolor{lightgray}
    \multicolumn{2}{l}{$\blacktriangledown$ \textit{InternVL3-9B}} && 66.3 & 13.2 & 53.0 && 49.9 & 65.3 & 75.4 & 53.8 \\
    & SFT && 66.2 & 16.9 & 51.7 && 53.2 & 69.7 & 74.8 & 55.4 \\
    & T-pref-DPO && 72.5 & 56.4 & 80.9 && 62.2 & 78.5 & \textbf{81.5} & 72.0 \\
    & PaMi-VDPO && 66.5 & 14.9 & 52.3 && 53.7 & 65.3 & 74.3 & 54.5 \\
    & \cellcolor{ourcolor}\textbf{\ours (Ours)}
      & \cellcolor{ourcolor}
      & \cellcolor{ourcolor}\textbf{74.1}
      & \cellcolor{ourcolor}\textbf{61.3}
      & \cellcolor{ourcolor}\textbf{84.7}
      & \cellcolor{ourcolor}
      & \cellcolor{ourcolor}\textbf{63.2}
      & \cellcolor{ourcolor}\textbf{79.6}
    & \cellcolor{ourcolor}81.1
    & \cellcolor{ourcolor}\textbf{74.0} \\
    \bottomrule
  \end{tabular}
  }
  \vspace{-0.1cm}
  \caption{\textbf{\dataset benchmark results.} Accuracy (\%) on temporal ordering and action recognition tasks for Qwen2.5-VL and InternVL3. FF: free-form; OL: order list; BC: binary choice; MC: multiple choice.}
  \label{tab:countervid}
  \vspace{-0.2cm}
\end{table}

\subsection{Experimental Results}

\begin{figure*}[t]
    \centering
    \includegraphics[width=0.99\linewidth]{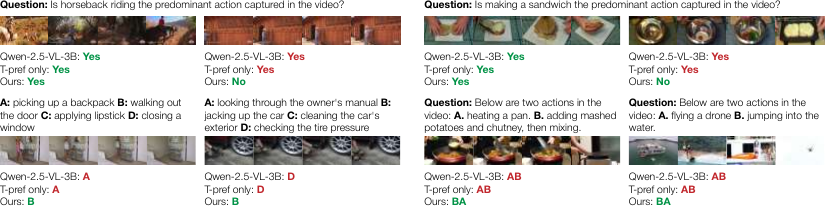}
    \vspace{-0.2cm}
    \caption{\textbf{Qualitative hallucination correction.} On VidHalluc, \ours produces responses that better follow visual evidence than the base and text-only preference models, particularly for temporal-sequence questions.}
    \label{fig:ex-results}
    \vspace{-0.3cm}
\end{figure*}

\tit{\dataset Benchmark}
Table~\ref{tab:countervid} reports results on the held-out \dataset benchmark across Qwen2.5-VL and InternVL3 models. \ours improves average accuracy from 36.9\% to 45.2\% on Qwen2.5-VL-3B, from 57.8\% to 66.2\% on Qwen2.5-VL-7B, and from 53.8\% to 74.0\% on InternVL3-9B. Gains are largest on temporal ordering, where predictions must track changed visual evidence: order-list accuracy rises from 1.6\% to 16.7\% for 3B and from 16.5\% to 43.8\% for 7B. On InternVL3-9B, \ours improves over T-pref-DPO by 2.0pp on average, with the largest gains on temporal ordering.

The SFT baseline yields minimal or negative changes (\eg, -0.7pp for 7B), while PaMi-VDPO struggles on \dataset, especially on temporal tasks, due to the lack of controlled action negatives.
Beyond answer-level preference learning, \ours provides the largest additional gains on temporal ordering: on Qwen2.5-VL-7B, it improves over T-pref-DPO by +6.7pp on order-list and +5.3pp on binary temporal questions. This indicates that visual preferences add input sensitivity when answer calibration alone is insufficient.

\begin{table}[t]
  \centering
  \setlength{\tabcolsep}{0.35em}
  \resizebox{\linewidth}{!}{
  \begin{tabular}{@{}p{0.6cm}@{}l c cccc}
    \toprule
    \multicolumn{2}{l}{} && \textbf{EventHal} & \textbf{VidHal} & \textbf{VideoHal} & \textbf{VideoHcr} \\
    \multicolumn{2}{l}{\textbf{Model}} && (Avg) & (TSH) & (Overall) & (Avg) \\
    \midrule
    \rowcolor{lightgray}
    \multicolumn{2}{l}{$\blacktriangledown$ \textit{Qwen2.5-VL-3B}} && 59.4 & 78.5 & 35.1 & 52.6 \\
    &SFT && 59.9 & \textbf{81.5} & 35.2 & 52.1 \\
    & T-pref-DPO && 64.3 & 80.0 & 36.3 & 53.6 \\
    & PaMi-VDPO && \textbf{66.4} & 76.2 & 36.3 & 50.7 \\
    & \cellcolor{ourcolor}\textbf{\ours (Ours)}
    & \cellcolor{ourcolor}
    & \cellcolor{ourcolor}64.4
    & \cellcolor{ourcolor}80.0
    & \cellcolor{ourcolor}\textbf{37.9}
    & \cellcolor{ourcolor}\textbf{54.6} \\
    \midrule
    \rowcolor{lightgray}
    \multicolumn{2}{l}{$\blacktriangledown$ \textit{Qwen2.5-VL-7B}} && 70.1 & 82.7 & 38.2 & 54.1 \\
    & SFT && 66.7 & 82.7 & 39.0 & 53.9 \\
    & T-pref-DPO && \textbf{72.7} & 87.8 & \textbf{39.8} & 55.1 \\
    & PaMi-VDPO && 68.3 & 83.0 & 38.7 & 55.1 \\
    & \cellcolor{ourcolor}\textbf{\ours (Ours)}
    & \cellcolor{ourcolor}
    & \cellcolor{ourcolor}72.5
    & \cellcolor{ourcolor}\textbf{90.5}
    & \cellcolor{ourcolor}39.3
    & \cellcolor{ourcolor}\textbf{55.7} \\
    \midrule
    \rowcolor{lightgray}
    \multicolumn{2}{l}{$\blacktriangledown$ \textit{InternVL3-9B}} && 64.6 & 39.3 & 26.5 & 54.6 \\
    & SFT && 64.0 & 46.5 & 27.4 & 55.6 \\
    & T-pref-DPO && 73.5 & 83.5 & 25.5 & \textbf{59.5} \\
    & PaMi-VDPO && 58.6 & 65.7 & \textbf{30.7} & 50.7 \\
    &\cellcolor{ourcolor}\textbf{\ours (Ours)}
    & \cellcolor{ourcolor}
    & \cellcolor{ourcolor}\textbf{73.7}
    & \cellcolor{ourcolor}\textbf{88.8}
    & \cellcolor{ourcolor}25.7
    & \cellcolor{ourcolor}58.9 \\
    \bottomrule
  \end{tabular}
  }
    \vspace{-0.1cm}
  \caption{\textbf{Performance on established hallucination benchmarks.} Accuracy (\%) across event-level, temporal, and content-based hallucination evaluations for Qwen2.5-VL and InternVL3 backbones.}
  \label{tab:hallucination}
  \vspace{-0.25cm}
\end{table}

\tit{Established Hallucination Benchmarks}
Table~\ref{tab:hallucination} reports results on four established hallucination benchmarks. \ours improves EventHallusion over the base model across all three backbones, reaching 64.4\%, 72.5\%, and 73.7\% on Qwen2.5-VL-3B, Qwen2.5-VL-7B, and InternVL3-9B, respectively. The clearest additional gains over T-pref-DPO appear on VidHalluc TSH, with +2.7pp on Qwen2.5-VL-7B and +5.3pp on InternVL3-9B, while the 3B result is tied. Overall, across the 12 matched external comparisons, \ours improves over T-pref-DPO in eight cases and ties in one. The three regressions are small: $-0.2$pp on EventHallusion and $-0.5$pp on VideoHallu for Qwen2.5-VL-7B, and $-0.6$pp on VideoHallucer for InternVL3-9B. Since these benchmarks differ from \dataset in scene composition, action distribution, and failure modes, these results support positive but non-uniform transfer beyond our synthetic benchmark, with the strongest gains on the temporal hallucination settings most directly targeted by our method.

\tit{General Video Understanding}
Table~\ref{tab:general} shows that general video understanding is largely preserved. VideoMME remains stable across Qwen2.5-VL and InternVL3, while NExT-QA improves for Qwen2.5-VL-3B and remains close to baseline for 7B and 9B. TempCompass improves for Qwen2.5-VL, including a caption-matching gain from 76.2\% to 80.2\% for 7B, but its caption-matching subset decreases for InternVL3. All benchmarks use their official splits without duration filtering; VideoMME therefore includes short, medium, and long videos. Although our method does not explicitly target long-form reasoning, evaluation on the general benchmarks tests its behavior beyond the short clips used for supervision.

\begin{table}[t]
  \centering
  \setlength{\tabcolsep}{0.35em}
  \resizebox{\linewidth}{!}{
  \begin{tabular}{@{}p{0.6cm}@{}l c cccccc}
    \toprule
    \multicolumn{2}{l}{} && \textbf{VideoMME} & \multicolumn{2}{c}{\textbf{NExT-QA}} & \multicolumn{3}{c}{\textbf{TempCompass}} \\
    \cmidrule(lr){4-4} \cmidrule(lr){5-6} \cmidrule(lr){7-9} 
    \multicolumn{2}{l}{\textbf{Model}} && (Acc) & MC & OE & MC & BC & CM \\
    \midrule
    \rowcolor{lightgray}
    \multicolumn{2}{l}{$\blacktriangledown$ \textit{Qwen2.5-VL-3B}} && \textbf{57.8} & 75.2 & 29.4 & 65.8 & 68.4 & 76.2 \\
    & SFT && 57.5 & 76.4 & 29.8 & 66.2 & 68.2 & 76.1 \\
    & T-pref-DPO && 57.4 & \textbf{76.8} & 29.6 & 66.8 & 69.0 & 77.5 \\
    & \cellcolor{ourcolor}\textbf{\ours (Ours)}
    & \cellcolor{ourcolor}
    & \cellcolor{ourcolor}\textbf{57.8}
    & \cellcolor{ourcolor}76.7
    & \cellcolor{ourcolor}\textbf{29.7}
    & \cellcolor{ourcolor}\textbf{67.0}
    & \cellcolor{ourcolor}\textbf{69.6}
    & \cellcolor{ourcolor}\textbf{77.2} \\
    \midrule
    \rowcolor{lightgray}
    \multicolumn{2}{l}{$\blacktriangledown$ \textit{Qwen2.5-VL-7B}} && 61.4 & 74.7 & \textbf{32.2} & 73.2 & 75.5 & 76.2 \\
    & SFT && \textbf{61.5} & \textbf{75.5} & \textbf{32.2} & 73.4 & 75.6 & 79.6 \\
    & T-pref-DPO && 61.4 & 75.1 & \textbf{32.2} & 73.4 & 75.8 & 77.6 \\
    & \cellcolor{ourcolor}\textbf{\ours (Ours)}
    & \cellcolor{ourcolor}
    & \cellcolor{ourcolor}\textbf{61.5}
    & \cellcolor{ourcolor}74.9
    & \cellcolor{ourcolor}\textbf{32.2}
    & \cellcolor{ourcolor}\textbf{73.7}
    & \cellcolor{ourcolor}\textbf{75.9}
    & \cellcolor{ourcolor}\textbf{80.2} \\
    \midrule
    \rowcolor{lightgray}
    \multicolumn{2}{l}{$\blacktriangledown$ \textit{InternVL3-9B}} && 64.8 & 80.7 & 33.2 & 69.3 & 70.9 & \textbf{76.7} \\
    & SFT && 64.5 & 80.8 & 33.0 & 69.8 & 72.4 & 76.4 \\
    & T-pref-DPO && \textbf{64.9} & \textbf{80.9} & \textbf{33.3} & \textbf{70.3} & 74.2 & 70.7 \\
    & \cellcolor{ourcolor}\textbf{\ours (Ours)}
    & \cellcolor{ourcolor}
    & \cellcolor{ourcolor}64.7
    & \cellcolor{ourcolor}\textbf{80.9}
    & \cellcolor{ourcolor}33.0
    & \cellcolor{ourcolor}70.1
    & \cellcolor{ourcolor}\textbf{74.4}
    & \cellcolor{ourcolor}69.5 \\
    \bottomrule
  \end{tabular}
  }
  \vspace{-0.1cm}
  \caption{\textbf{General video benchmarks.} Accuracy (\%) on VideoMME, NExT-QA, and TempCompass for Qwen2.5-VL and InternVL3. MC: multiple choice; OE: open-ended; BC: binary choice; CM: caption matching.}
  \label{tab:general}
\vspace{-0.25cm}
\end{table}

\subsection{Qualitative Analysis}
Fig.~\ref{fig:ex-results} presents representative VidHalluc examples. The base and text-only preference models often favor plausible but unsupported actions or event sequences, whereas \ours produces responses that better follow the video content and correct both action and temporal hallucinations. Beyond correcting explicit errors, we observe that \ours exhibits more cautious behavior when the visual evidence is ambiguous or partially occluded, avoiding confident but unsupported claims. This qualitative behavior complements the quantitative results, suggesting that mixed visual and textual preference learning improves both visual grounding and sensitivity to action order.

\subsection{Robustness Analysis}
\label{sec:robustness}

Table~\ref{tab:sensitivity_compact} shows that \ours is not sensitive to a narrow balance between preference types. Changing $\lambda$ fourfold alters average accuracy by at most 0.4pp on Qwen2.5-VL-7B and 0.2pp on InternVL3-9B. Moving from a 50/50 to 90/10 visual/textual mixture changes the average by 0.5pp and 1.0pp, respectively. Although 90/10 is highest overall, it exceeds 70/30 by only 0.4pp on each backbone and does not improve every task group. We treat the sweep as a robustness analysis rather than selecting a new mixture on the same evaluation benchmark, retaining 70/30 as the balanced configuration used throughout. Appendix~\ref{sec:sensitivity_details} reports all six submetrics and the complementary $\beta$ sweep.

\begin{table}[t]
  \centering
  \setlength{\tabcolsep}{0.45em}
  \resizebox{\linewidth}{!}{
  \begin{tabular}{llccc}
    \toprule
    \textbf{Model} & \textbf{Setting} & \textbf{TempOrd} & \textbf{ActRec} & \textbf{Avg.} \\
    \midrule
    \multirow{5}{*}{Qwen2.5-VL-7B}
      & $\lambda{=}1$, 70/30 & 62.9 & 69.3 & 66.1 \\
      & $\lambda{=}0.5$ & 62.9 & 69.0 & 66.0 \\
      & $\lambda{=}2.0$ & 62.3 & 69.1 & 65.7 \\
      & v/t $=50/50$ & 62.7 & 69.3 & 66.0 \\
      & v/t $=90/10$ & \textbf{64.2} & 68.8 & \textbf{66.5} \\
    \midrule
    \multirow{5}{*}{InternVL3-9B}
      & $\lambda{=}1$, 70/30 & 73.2 & 74.7 & 74.0 \\
      & $\lambda{=}0.5$ & 73.1 & \textbf{74.9} & 74.0 \\
      & $\lambda{=}2.0$ & 73.0 & 74.6 & 73.8 \\
      & v/t $=50/50$ & 72.0 & 74.8 & 73.4 \\
      & v/t $=90/10$ & \textbf{74.1} & 74.7 & \textbf{74.4} \\
    \bottomrule
  \end{tabular}
  }
  \vspace{-0.15cm}
  \caption{\textbf{Loss-weight and preference-mixture sensitivity.} Controlled \dataset configurations varying one factor around $\lambda{=}1$ and v/t $=70/30$.}
  \label{tab:sensitivity_compact}
  \vspace{-0.3cm}
\end{table}

\section{Conclusion}
We presented a scalable framework for mitigating hallucinations in VLMs by coupling counterfactual video generation with unified preference-based alignment. By synthesizing semantic hard negatives that preserve scene context while varying action dynamics, our approach directly targets action misidentification and temporal-ordering errors that commonly arise from over-reliance on language priors. Across Qwen2.5-VL and InternVL3 backbones, \dataset and \ours yield their strongest gains on these targeted capabilities, improve most external hallucination benchmark comparisons, and largely preserve general video understanding. This work highlights the promise of generative counterfactual data as a practical mechanism for improving visual grounding in VLMs, while leaving long-form temporal reasoning and broader component replacement for future work.

\section*{Limitations}
While our approach achieves strong performance, several aspects offer opportunities for future extension and refinement. First, the evaluation mechanisms used for action proposal filtering and video generation could be further refined, for example, through more comprehensive sanity checks, to increase the proportion of high-quality samples in the final dataset. Second, the quality of the generated counterfactuals depends on the underlying generative models and is expected to improve as image and video generation methods continue to advance. Residual synthesis artifacts may also provide shortcut signals; our quality controls and external transfer results make a purely artifact-driven explanation less likely, but do not eliminate this confound. Third, our current counterfactuals focus on short actions ($\approx 3\,\mathrm{s}$) and controlled concatenations of those clips. Although we evaluate official external splits without duration filtering, including VideoMME's short, medium, and long videos, the method does not directly address long-range causal dependencies, overlapping events, or multi-agent interactions.
Fourth, the pipeline is modular at the interface level and alternative VLM judges reproduce most filtering decisions, but we do not establish invariant quality under replacement of the image editor or video generator; cross-generator validation remains future work.
Finally, we adopt parameter-efficient fine-tuning by freezing the vision encoder; while full end-to-end training may further enhance sensitivity to fine-grained visual cues, it incurs substantially higher computational cost and optimization complexity, which we leave for future exploration.

\section*{Ethical Considerations}
The ability to synthesize context-aware alternative actions via our counterfactual generation framework is a dual-use technology; while designed for model alignment, it could potentially be repurposed to create misinformation or deceptive video content. We acknowledge that releasing our methodology and code inherently carries this risk, yet we prioritize the scientific necessity of addressing the widespread problem of VLM hallucinations. In addition, because our framework relies on large-scale generative models, the resulting synthetic data may inherit biases or stereotypes present in those models. To reduce harmful content in the released dataset, we screen generated videos with automated visual safety classifiers (NudeNet~\cite{bedapudi2019nudenet} and Q16~\cite{schramowski2022can}) and remove the approximately 0.6\% of samples flagged as potentially unsafe. The dataset is derived from PVD only through anchor frames; it does not include source metadata and follows the privacy protections of the original source data. Finally, we note that large-scale synthetic video generation has environmental costs; by providing our code to the community, we enable other researchers to build upon our findings and alignment methodology without the need for redundant, large-scale computational trials.

\section*{Acknowledgments}

Marcella Cornia, Lorenzo Baraldi, and Rita Cucchiara acknowledge funding from the EU Horizon project ELLIOT (GA No. 101214398).

\bibliography{bibliography}

\appendix

\newpage
\section{Additional Implementation and Training Details}
\label{sec:training-details-addl}

\subsection{Data Generation}
\tinytit{Detailed Data Generation Configuration}
All stages of the counterfactual video generation pipeline are executed in a modular and parallelized manner, with keyframe extraction, action proposal, image editing, and video synthesis run independently across up to three NVIDIA A100 GPUs (40GB). Generating the $\sim$26k preference pairs in \dataset required approximately 3{,}000 GPU-hours and about \$60 in LLM API cost for proposal, filtering, and refinement. Keyframe extraction uses SigLIP-SO400M\footnote{\href{https://huggingface.co/google/siglip-so400m-patch14-384}{\texttt{google/siglip-so400m-patch14-384}}} for caption-frame alignment, sampling frames with a temporal stride of $0.5$s and averaging similarity over multiple spatial crops. Action proposal and filtering are performed with Claude-4-Sonnet using a temperature equal to $0.0$ for deterministic outputs. End-frame generation is conducted with Qwen-Image-Edit\footnote{\href{https://huggingface.co/Qwen/Qwen-Image-Edit}{\texttt{Qwen/Qwen-Image-Edit}}} using the Lightning-4steps-V1.0 LoRA adapter\footnote{\href{https://huggingface.co/lightx2v/Qwen-Image-Lightning}{\texttt{lightx2v/Qwen-Image-Lightning}}}, running in FP8 precision with four diffusion steps, classifier-free guidance equal to 1.0, and denoising strength 0.9. An iterative refinement loop with a maximum of five attempts is used to ensure action correctness and scene consistency. Video synthesis employs Wan2.2-I2V-14B\footnote{\href{https://huggingface.co/Wan-AI/Wan2.2-I2V-A14B}{\texttt{Wan-AI/Wan2.2-I2V-A14B}}} in FP8 precision with the I2V\_lightx2v\_4steps LoRA adapter\footnote{\href{https://huggingface.co/Comfy-Org/Wan_2.2_ComfyUI_Repackaged}{\texttt{Comfy-Org/Wan\_2.2\_ComfyUI\_Repackaged}}}, generating videos at $680{\times}384$ resolution with 49 frames at 16 fps, corresponding to a duration of approximately 3.06 seconds.

\tit{Safety Filtering}
We apply automated visual safety screening to the generated samples using NudeNet~\cite{bedapudi2019nudenet}, targeting sexual content, and Q16~\cite{schramowski2022can}, covering broader harmful categories such as gore and violence. These classifiers flag approximately 0.6\% of generated samples as potentially unsafe; flagged samples are removed from the final released dataset. Because \dataset is derived from PVD only through anchor frames, it does not include source metadata, and we inherit PVD's privacy-protection procedures for the real anchors.

\tit{Human Annotation Protocol}
Human evaluators use an interface that displays the visual input, the prompt, and the target answer for each sampled preference item. For output-swap examples, the interface shows one visual input and two answers, and annotators evaluate whether the chosen answer is correct for that visual input. For input-swap examples, it shows the chosen and rejected visual inputs with a single answer, and annotators evaluate the answer only with respect to the chosen visual input. Separately, annotators rate the visual quality of the visual input, independently of answer correctness. This two-axis protocol prevents the ``good'' label from conflating semantic validity with visual fidelity and makes label noise and generation artifacts separately observable.

\subsection{Training and Evaluation}
\tinytit{Training Configuration}
To train \ours, we fine-tune Qwen2.5-VL models (3B\footnote{\href{https://huggingface.co/Qwen/Qwen2.5-VL-3B-Instruct}{\texttt{Qwen/Qwen2.5-VL-3B-Instruct}}} and 7B\footnote{\href{https://huggingface.co/Qwen/Qwen2.5-VL-7B-Instruct}{\texttt{Qwen/Qwen2.5-VL-7B-Instruct}}}) and InternVL3-9B\footnote{\href{https://huggingface.co/OpenGVLab/InternVL3-9B}{\texttt{OpenGVLab/InternVL3-9B}}} using parameter-efficient adaptation. The vision encoder is frozen except for the vision-language merger module, while the language model is adapted with LoRA~\cite{hu2022lora} applied to attention layers (rank 64, $\alpha{=}16$). Training is performed for three epochs at 2 fps with a learning rate of $1{\times}10^{-6}$ using the AdamW optimizer~\cite{loshchilov2017decoupled}, an effective batch size of 8 distributed over 8 GPUs, and a KL penalty coefficient of $\beta{=}0.7$. The reference policy $\pi_{\text{ref}}$ is a frozen copy of the corresponding initial checkpoint.

\tit{Evaluation Configuration}
For reproducibility, we apply a consistent pre-processing setup across all evaluation benchmarks. For all image and video inputs, we enforce a strict pixel budget where the maximum resolution is set to 151,200 pixels. Similarly, the minimum resolution is maintained at 100,352 pixels. Regarding temporal sampling, the model is limited to a maximum of 32 frames for all video-based benchmarks.

\tit{PaMi-VDPO Baseline Configuration}
For the PaMi-VDPO baseline~\cite{ding2025pami_vdpo}, we construct two augmentation-based rejected instances for each chosen video. Specifically, we apply \textit{Crop}, which randomly crops at most 20\% of the spatial region and resizes the result back to the original resolution, and \textit{Shuffle}, which randomly permutes the temporal order of frames or short segments. The chosen video and both rejected instances form a single multi-instance training example. The PaMi-VDPO objective derives prompt/model-dependent weights from token-level divergence scores and jointly applies them to the two negative rewards. Both variants use the same three-epoch budget and backbone-specific optimization settings as the matched comparison. In the main tables, the chosen videos are \dataset-generated clips; Appendix~\ref{sec:supp_results} reports both the \dataset- and PVD-anchor variants.

\begin{table}[t]
  \centering
  \setlength{\tabcolsep}{0.35em}
  \resizebox{\linewidth}{!}{
  \begin{tabular}{l ccc ccc c}
    \toprule
    \textbf{Model} & \multicolumn{3}{c}{\textbf{Temporal Ord.}} & \multicolumn{3}{c}{\textbf{Action Rec.}} & \textbf{Avg} \\
    & FF & OL & BC & FF & MC & BC & \\
    \midrule
    \rowcolor{lightgray}
    $\blacktriangledown$ \textit{Qwen2.5-VL-3B} & 29.8 & 1.6 & 48.9 & 28.0 & 48.4 & 64.6 & 36.9 \\
    PaMi-VDPO (PVD)   & 32.2 & 1.6 & 47.1 & 29.1 & 51.4 & 67.6 & 38.2 \\
    PaMi-VDPO (CtVid) & 32.9 & 1.4 & 47.5 & 29.1 & 52.3 & 68.0 & 38.5 \\
    \cellcolor{ourcolor}\textbf{\ours (Ours)}
      & \cellcolor{ourcolor}31.0
      & \cellcolor{ourcolor}16.7
      & \cellcolor{ourcolor}64.0
      & \cellcolor{ourcolor}29.5
      & \cellcolor{ourcolor}60.0
      & \cellcolor{ourcolor}70.3
      & \cellcolor{ourcolor}45.2 \\
    \midrule
    \rowcolor{lightgray}
    $\blacktriangledown$ \textit{Qwen2.5-VL-7B} & 70.3 & 16.5 & 57.8 & 55.1 & 68.6 & 78.2 & 57.8 \\
    PaMi-VDPO (PVD)   & 42.3 & 13.2 & 57.6 & 33.0 & 64.2 & 74.1 & 47.4 \\
    PaMi-VDPO (CtVid) & 42.2 & 13.4 & 57.6 & 32.9 & 64.4 & 74.5 & 47.5 \\
    \cellcolor{ourcolor}\textbf{\ours (Ours)}
      & \cellcolor{ourcolor}72.2
      & \cellcolor{ourcolor}43.8
      & \cellcolor{ourcolor}72.7
      & \cellcolor{ourcolor}56.3
      & \cellcolor{ourcolor}71.2
      & \cellcolor{ourcolor}80.7
      & \cellcolor{ourcolor}66.2 \\
    \midrule
    \rowcolor{lightgray}
    $\blacktriangledown$ \textit{InternVL3-9B} & 66.3 & 13.2 & 53.0 & 49.9 & 65.3 & 75.4 & 53.8 \\
    PaMi-VDPO (PVD)   & 65.4 & 12.8 & 56.0 & 50.4 & 64.4 & 75.8 & 54.1 \\
    PaMi-VDPO (CtVid) & 66.5 & 14.9 & 52.3 & 53.7 & 65.3 & 74.3 & 54.5 \\
    \cellcolor{ourcolor}\textbf{\ours (Ours)}
      & \cellcolor{ourcolor}74.1
      & \cellcolor{ourcolor}61.3
      & \cellcolor{ourcolor}84.7
      & \cellcolor{ourcolor}63.2
      & \cellcolor{ourcolor}79.6
      & \cellcolor{ourcolor}81.1
      & \cellcolor{ourcolor}74.0 \\
    \bottomrule
  \end{tabular}
  }
  \vspace{-0.15cm}
  \caption{\textbf{PaMi-VDPO variant breakdown on \dataset.} We compare the PVD-anchor and \dataset-anchor variants of the PaMi-VDPO baseline under the same augmentation-based negatives, and report the corresponding base and \ours results for context.}
  \label{tab:pami_variants}
  \vspace{-0.15cm}
\end{table}

\begin{table}[t]
  \centering
  \setlength{\tabcolsep}{0.65em}
  \resizebox{\linewidth}{!}{
  \begin{tabular}{lcc}
    \toprule
    \textbf{Samples} & \textbf{CLIP-Score} & \textbf{VQAScore} \\
    \midrule
    Real PVD pairs & 0.268 & 0.720 \\
    Generated \dataset pairs & 0.251 & 0.673 \\
    \bottomrule
  \end{tabular}
  }
  \vspace{-0.15cm}
  \caption{\textbf{Automatic semantic quality metrics.} We compare original PVD video-caption pairs and generated \dataset pairs under the same CLIP-Score~\cite{hessel2021clipscore} and VQAScore~\cite{lin2024evaluating} protocol.}
  \label{tab:generation_quality_metrics}
  \vspace{-0.3cm}
\end{table}

\begin{table*}[t]
  \centering
  \footnotesize
  \setlength{\tabcolsep}{0.6em}
  \renewcommand{\arraystretch}{0.95}
  \begin{tabular}{ll ccc ccc c}
    \toprule
    \multicolumn{2}{l}{} & \multicolumn{3}{c}{\textbf{Temporal Ord.}} & \multicolumn{3}{c}{\textbf{Action Rec.}} & \\
    \cmidrule(lr){3-5} \cmidrule(lr){6-8}
    \textbf{Model} & \textbf{Setting} & FF & OL & BC & FF & MC & BC & \textbf{Avg.} \\
    \midrule
    \multirow{7}{*}{Qwen2.5-VL-7B}
      & Default: $\lambda{=}1,\beta{=}0.7$, 70/30 & 72.2 & 43.8 & 72.7 & 56.3 & 71.2 & 80.7 & 66.2 \\
      & $\lambda{=}0.5$ & 72.2 & 43.8 & 72.7 & 56.2 & 70.5 & 80.4 & 66.0 \\
      & $\lambda{=}2.0$ & 72.3 & 42.2 & 72.5 & 56.2 & 71.0 & 80.2 & 65.7 \\
      & $\beta{=}0.5$ & 72.4 & 51.1 & 77.4 & 56.6 & 69.0 & 81.5 & 68.0 \\
      & $\beta{=}1.0$ & 71.6 & 35.4 & 68.2 & 56.0 & 71.0 & 80.4 & 63.8 \\
      & v/t $=50/50$ & 72.3 & 42.8 & 72.9 & 56.3 & 71.0 & 80.7 & 66.0 \\
      & v/t $=90/10$ & 72.1 & 46.4 & 74.1 & 56.3 & 70.1 & 80.0 & 66.5 \\
    \midrule
    \multirow{7}{*}{InternVL3-9B}
      & Default: $\lambda{=}1,\beta{=}0.7$, 70/30 & 74.1 & 61.3 & 84.7 & 63.2 & 79.6 & 81.1 & 74.0 \\
      & $\lambda{=}0.5$ & 74.2 & 61.1 & 84.1 & 63.0 & 79.8 & 81.9 & 74.0 \\
      & $\lambda{=}2.0$ & 74.0 & 60.5 & 84.5 & 63.1 & 79.3 & 81.3 & 73.8 \\
      & $\beta{=}0.5$ & 74.8 & 62.5 & 84.9 & 63.5 & 79.6 & 82.1 & 74.6 \\
      & $\beta{=}1.0$ & 73.0 & 58.0 & 82.1 & 62.7 & 79.8 & 80.9 & 72.7 \\
      & v/t $=50/50$ & 73.7 & 59.5 & 82.7 & 62.6 & 79.6 & 82.1 & 73.4 \\
      & v/t $=90/10$ & 74.6 & 61.9 & 85.7 & 62.9 & 79.3 & 81.9 & 74.4 \\
    \bottomrule
  \end{tabular}
  \vspace{-0.15cm}
  \caption{\textbf{Hyperparameter and preference-composition sensitivity on \dataset.} Each block reports a matched default configuration and changes only the indicated factor. 70/30 denotes the visual/textual preference ratio.}
  \label{tab:sensitivity_full}
  \vspace{-0.4cm}
\end{table*}

\section{Additional Results and Analyses}
\label{sec:supp_results}
\tinytit{Detailed Comparison with PaMi-VDPO}
In Table~\ref{tab:pami_variants}, we separate two controlled variants that use the same prompt-aware multi-instance objective and augmentation-based rejected instances, but differ in the source of the chosen videos. In \textbf{PaMi-VDPO (PVD)}, the chosen videos are original PVD clips; in \textbf{PaMi-VDPO (CtVid)}, they are \dataset-generated clips, matching the chosen-video distribution used by \ours in the main paper. The CtVid variant therefore controls for the chosen-video source while replacing our semantic counterfactual negatives with generic Crop and Shuffle perturbations. In both variants, the two negatives are processed jointly rather than as independent DPO pairs.

\tit{Cross-Model Agreement}
To assess dependence on a single closed-source evaluator, we evaluate the two Claude-based filtering stages (action proposal filtering and edited-frame evaluation) on 5{,}000 samples using two open-source VLM judges, Qwen2.5-VL-7B and InternVL3-14B. The accept/reject decisions agree with Claude in 93.6\% of cases for Qwen2.5-VL-7B and 76.3\% for InternVL3-14B. This suggests that the filtering criteria are not unique to Claude. It does not, however, establish full interchangeability of the image editor or video generator, which we leave for future cross-generator validation.

\tit{Automatic Semantic Quality Metrics}
Table~\ref{tab:generation_quality_metrics} complements the human study shown in Table~\ref{tab:human_eval_tasks} with automatic semantic metrics. Specifically, CLIP-Score~\cite{hessel2021clipscore} measures cross-modal video-caption alignment, while VQAScore~\cite{lin2024evaluating} estimates answer consistency through visual question answering. Under the same evaluation protocol, generated \dataset pairs remain close to the original PVD pairs (0.251 vs.\ 0.268 CLIP-Score and 0.673 vs.\ 0.720 VQAScore), suggesting that the pipeline largely preserves semantic alignment while introducing controlled action or temporal variations.

\subsection{Hyperparameter Sensitivity}
\label{sec:sensitivity_details}

We vary one factor at a time around the paper configuration, using the same source pool, train/validation partition, total number of examples, and \dataset evaluation protocol. Table~\ref{tab:sensitivity_full} reports the six task submetrics for the visual-preference loss weight $\lambda$, the visual/textual preference mixture, and the DPO coefficient $\beta$.

The $\lambda$ sweep confirms stability across a fourfold range: average accuracy changes by at most 0.5pp on Qwen2.5-VL-7B and 0.2pp on InternVL3-9B. Preference composition is similarly stable. Although 90/10 improves the average by 0.3pp and 0.4pp, respectively, the gains are not consistent across submetrics; we therefore retain the 70/30 setting rather than selecting for a marginal advantage on this benchmark. The $\beta$ sweep has a larger effect: $\beta{=}0.5$ improves the average by 1.8pp and 0.6pp, whereas $\beta{=}1.0$ lowers it by 2.4pp and 1.3pp. Because these variants were evaluated only on \dataset, we retain $\beta{=}0.7$ as a conservative default.

\begin{table}[t]
  \centering
  \setlength{\tabcolsep}{0.35em}
  \resizebox{\linewidth}{!}{
  \begin{tabular}{@{}p{0.6cm}@{}l c ccc c ccc c}
    \toprule
    \multicolumn{2}{l}{} && \multicolumn{3}{c}{\textbf{Temporal Ord.}} && \multicolumn{3}{c}{\textbf{Action Rec.}} & \\
    \cmidrule{4-6} \cmidrule{8-10}
    \multicolumn{2}{l}{\textbf{Model}} && FF & OL & BC && FF & MC & BC & \textbf{Avg} \\
    \midrule
    \rowcolor{lightgray}
    \multicolumn{2}{l}{$\blacktriangledown$ \textit{InternVL3-9B}} && 66.3 & 13.2 & 53.0 && 49.9 & 65.3 & 75.4 & 53.8 \\
    & T-pref-DPO && 72.5 & 56.4 & 80.9 && 62.2 & 78.5 & \textbf{81.5} & 72.0 \\
    & \cellcolor{ourcolor}\textbf{\ours (Ours)}
      & \cellcolor{ourcolor}
      & \cellcolor{ourcolor}\textbf{74.1}
      & \cellcolor{ourcolor}\textbf{61.3}
      & \cellcolor{ourcolor}\textbf{84.7}
      & \cellcolor{ourcolor}
      & \cellcolor{ourcolor}\textbf{63.2}
      & \cellcolor{ourcolor}\textbf{79.6}
      & \cellcolor{ourcolor}81.1
      & \cellcolor{ourcolor}\textbf{74.0} \\
    \midrule
    \rowcolor{lightgray}
    \multicolumn{2}{l}{$\blacktriangledown$ \textit{InternVL3-14B}} && 68.3 & 15.9 & 63.3 && 54.4 & 66.8 & 72.7 & 56.9 \\
    & T-pref-DPO && 76.8 & 66.0 & 91.9 && 64.7 & 82.2 & 82.7 & 77.4 \\
    & \cellcolor{ourcolor}\textbf{\ours (Ours)}
      & \cellcolor{ourcolor}
      & \cellcolor{ourcolor}\textbf{77.6}
      & \cellcolor{ourcolor}\textbf{66.4}
      & \cellcolor{ourcolor}\textbf{92.3}
      & \cellcolor{ourcolor}
      & \cellcolor{ourcolor}\textbf{65.5}
      & \cellcolor{ourcolor}\textbf{82.6}
      & \cellcolor{ourcolor}\textbf{83.5}
      & \cellcolor{ourcolor}\textbf{78.0} \\
    \bottomrule
  \end{tabular}
  }
  \vspace{-0.15cm}
  \caption{\textbf{Scaling experiment on \dataset.} Accuracy (\%) for InternVL3-9B and InternVL3-14B under the matched Base, textual-preference DPO, and \ours configurations. FF: free-form; OL: order list; BC: binary choice; MC: multiple choice.}
  \label{tab:internvl_scale}
  \vspace{-0.3cm}
\end{table}

\subsection{Scaling to InternVL3-14B}
\label{sec:scale_14b}

To examine whether the behavior observed in the main experiments extends to a larger model, we train InternVL3-14B using the same \dataset data, \ours objective, and training protocol. Table~\ref{tab:internvl_scale} reports this experiment together with InternVL3-9B as a matched reference, comparing only Base, T-pref-DPO, and \ours on \dataset. At 14B, \ours improves over Base by 21.1 points on average and over T-pref-DPO by 0.6 points, with gains on all six submetrics. This provides evidence that visual-preference supervision remains useful at this scale, while not establishing behavior at frontier-scale models.

\subsection{Synthetic-Data Failure Analysis}
\label{sec:failure_analysis}

\begin{figure*}[t]
  \centering
  \includegraphics[width=0.98\textwidth]{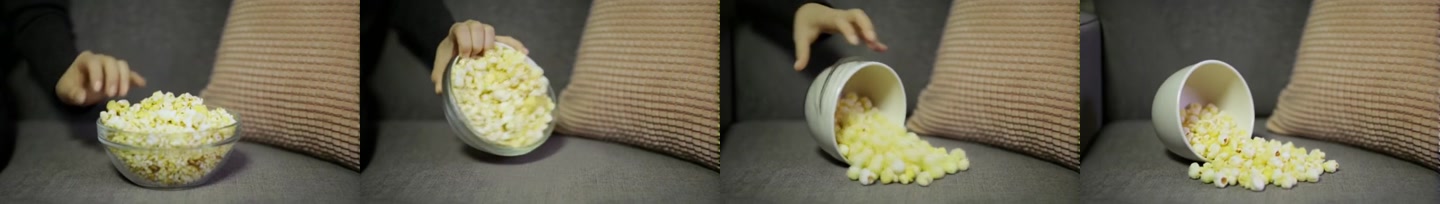}\\[-1pt]
  \small (a) \textcolor{green!50!black}{\textbf{Successful action realization.}} Target: tip the bowl and spill the popcorn.\par\smallskip
  \includegraphics[width=0.98\textwidth]{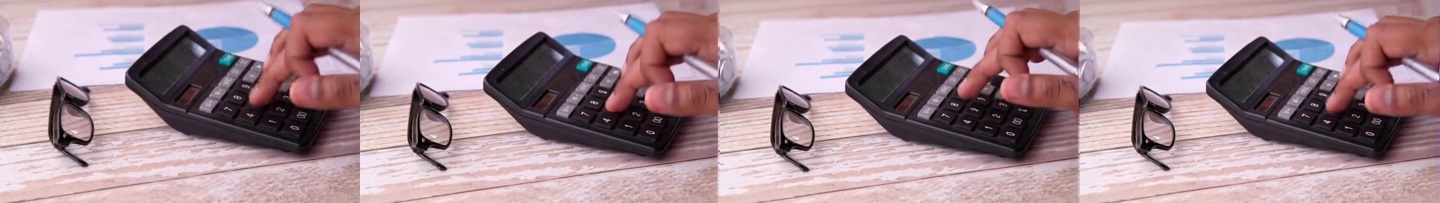}\\[-1pt]
  \small (b) \textcolor{red!75!black}{\textbf{Failed action realization.}} Target: slide the graph paper; it remains fixed.\par\smallskip
  \includegraphics[width=0.98\textwidth]{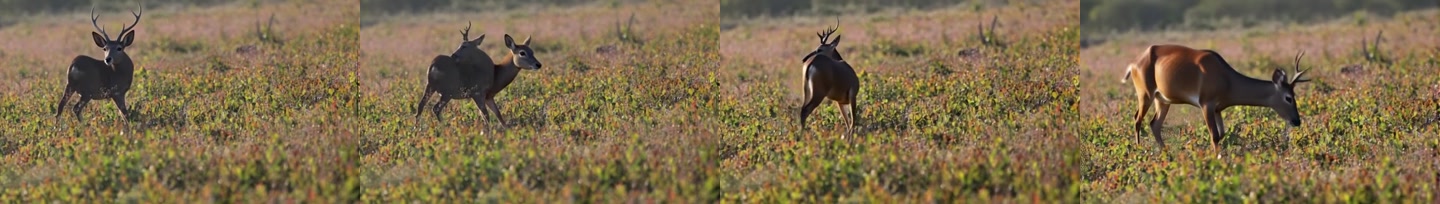}\\[-1pt]
  \small (c) \textcolor{red!75!black}{\textbf{Generation artifacts.}} Targets: perk ears, step, leap, turn, graze; identity/anatomy drift.\par\smallskip
  \includegraphics[width=0.98\textwidth]{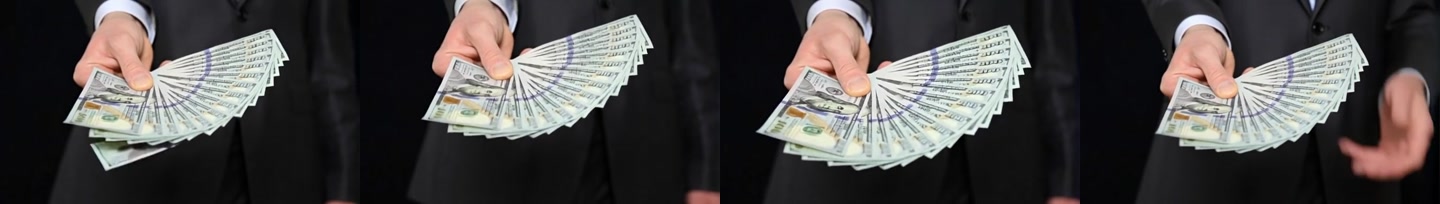}\\[-1pt]
  \small (d) \textcolor{red!75!black}{\textbf{Ambiguous action label.}} Targets: drop bills, fan wider, retract hand; actions and order are unclear.\par
  \vspace{-0.1cm}
  \caption{\textbf{Representative successful generation and failure cases.} Each row shows four uniformly spaced frames from one generated clip or concatenated action sequence. The examples make action realization and residual synthesis errors directly inspectable; the aggregate failure distribution is reported in Table~\ref{tab:failure_breakdown}.}
  \label{fig:generation_success_failures}
  \vspace{-0.3cm}
\end{figure*}

Following the human study in Sec.~\ref{subsec:human_eval_new}, we categorize the 46 unique samples flagged for an incorrect or ambiguous label, poor visual quality, or both. Table~\ref{tab:failure_breakdown} reports the resulting failure modes. Categories are not mutually exclusive, so counts and percentages need not sum to 46 or 100\%. Failed action realization is the dominant residual issue, followed by visible generation artifacts and ambiguous action labels. We observe no scene-identity drift or temporal-concatenation failures in this set. Fig.~\ref{fig:generation_success_failures} complements this aggregate analysis with a successful generation and representative cases from all three nonzero failure categories. The examples distinguish clean action realization from an unchanged intended state, visible synthesis inconsistencies, and action descriptions that are not visually separable.

\begin{table}[t]
  \centering
  \setlength{\tabcolsep}{0.65em}
  \resizebox{\linewidth}{!}{%
  \begin{tabular}{lrr}
    \toprule
    \textbf{Failure category} & \textbf{Count} & \textbf{Errors (\%)} \\
    \midrule
    Failed action realization & 39 & 84.8 \\
    Generation artifacts & 18 & 39.1 \\
    Ambiguous action label & 11 & 23.9 \\
    Scene identity drift & 0 & 0.0 \\
    Temporal concatenation artifacts & 0 & 0.0 \\
    \bottomrule
  \end{tabular}
  }
  \vspace{-0.15cm}
  \caption{\textbf{Failure-mode breakdown for the 46 samples flagged during human evaluation.} Flags indicate an incorrect or ambiguous label, poor visual quality, or both; multiple failure categories may apply per sample.}
  \label{tab:failure_breakdown}
    \vspace{-0.5cm}
\end{table}

\section{Evaluation Benchmarks Details}
\label{sec:benchmark-details}
To provide a comprehensive evaluation of model performance, we employ a diverse set of benchmarks, targeting both specific hallucination modes and general video understanding capabilities.

\subsection{Video Hallucination Benchmarks}
We evaluate on four benchmarks designed to assess complementary aspects of multimodal hallucination, ranging from object existence to event-level and temporal reasoning.

\tit{VideoHallucer~\cite{wang2024videohallucer}}
It comprises approximately 1,000 samples evaluated via binary-choice question answering. It distinguishes between intrinsic hallucinations, corresponding to incorrect descriptions of content present in the video (\eg, object relations, temporal sequencing, semantic details), and extrinsic hallucinations, which involve fabricating content absent from the video, often driven by language priors, including factual errors and unsupported inventions.

\tit{VidHalluc~\cite{im2025vidhalluc}}
VidHalluc is a large-scale benchmark containing approximately 5,000 videos and 9,000 question-answer pairs.
It evaluates models across binary-choice, multiple-choice, sorting, and open-ended formats, targeting action hallucination (misidentification of actions), temporal sequence hallucination (errors in event ordering), and scene transition hallucination (misinterpretation of changes across scenes). In our experiments, we report results on the temporal sequence hallucination (TSH) samples.

\tit{VideoHallu~\cite{li2025videohallu}}
This benchmark consists of approximately 3{,}000 synthetic videos designed for short-answer question answering. It evaluates hallucinations across alignment (entity counting, recognition, and entity properties), spatial-temporal consistency (camera, spatial, and temporal dynamics), physics reasoning (constraints, properties, motion, and state transitions), and commonsense (general knowledge and reasoning capabilities).

\tit{EventHallusion~\cite{eventhallusion2024}}
EventHallusion comprises 400 videos and about 700 question-answer pairs, focusing on event-level hallucinations driven by parametric knowledge (language priors). It categorizes videos into three types: \textit{entire rare events}, which depict uncommon events to test reliance on priors; \textit{interleave}, which features a mix of common and rare events occurring sequentially; and \textit{misleading}, which pairs videos of common events with customized questions designed to trigger misleading event demonstrations.

\subsection{General Video Understanding Benchmarks}
To ensure that mitigating hallucinations does not compromise general capabilities, we also evaluate on three standard benchmarks.

\tit{VideoMME~\cite{fu2024videomme}}
VideoMME is a comprehensive benchmark for multimodal video understanding that assesses the model's ability to process and reason over diverse video domains.

\tit{NExT-QA~\cite{xiao2021nextqa}}
This benchmark is specifically designed to evaluate causal and temporal action reasoning. It challenges models to explain ``why'' and ``how'' actions occur, moving beyond simple descriptive tasks.

\tit{TempCompass~\cite{liu2024tempcompass}}
Focused on temporal commonsense reasoning, it employs multiple-choice, binary-choice, and caption matching formats to test understanding of temporal concepts such as duration, frequency, and event order.

\section{Generation Pipeline Prompts}
\label{sec:prompts}
Our counterfactual generation pipeline is implemented through a sequence of structured prompts that govern action proposal, action filtering, image editing, and iterative refinement. For completeness and reproducibility, we report the system prompts used in each stage of the pipeline.

\tit{Action Proposal and Filtering}
Conditioned on the anchor frame $I^{\text{start}}$ and the original caption $C^{\text{real}}$, a multimodal LLM is first prompted to propose $N$ alternative action captions $\{C_i\}_{i=1}^N$.
As shown in Fig.~\ref{fig:prompt_a}, the action proposal prompt enforces that each action is contextually appropriate, involves visible subjects or objects, is achievable via image editing, and is semantically distinct from the others. To remove implausible or redundant proposals, a second prompt evaluates each action according to subject presence, physical and contextual feasibility, and semantic uniqueness (Fig.~\ref{fig:prompt_b}).
Actions that fail any criterion (such as relying on absent objects or being minor variants of another action) are discarded, ensuring a diverse set of visually expressible hard negatives.

\tit{Editing Instruction Generation and Refinement}
For each retained action caption $C_i$, we prompt the same model to generate a structured image-editing instruction that transforms the anchor frame into a plausible final-state depiction of the action.
As illustrated in Fig.~\ref{fig:prompt_c}, this prompt constrains the instruction to preserve scene identity, camera parameters, and unrelated content, while explicitly describing the visible outcome of the action. 

Because image editing may introduce semantic drift or artifacts, we employ an iterative verification loop. After each edit, an evaluation prompt compares the original and edited frames to determine whether the intended action has been correctly realized while maintaining scene consistency (Fig.~\ref{fig:prompt_d}). If the edit fails, a refinement prompt generates a new instruction explicitly differentiated from all previous attempts and targets the observed failure modes (Fig.~\ref{fig:prompt_e}).
This refinement process is repeated up to a fixed number of attempts, yielding a validated end frame for subsequent video synthesis.

\begin{figure*}[t]
    \centering
    \begin{tcolorbox}[
        colback=promptbg,      
        colframe=promptborder, 
        arc=2mm,               
        boxrule=1pt,           
        width=\linewidth,      
        left=10pt, right=10pt, top=10pt, bottom=10pt
    ]
        \ttfamily\small 
You are given a single video frame (image) and the original video caption.

\textbf{ORIGINAL CONTEXT (caption):} \placeholder{caption}

\textbf{TASK:} Propose EXACTLY \placeholder{num\_actions} simple, realistic actions that could plausibly happen in THIS scene. Each action must be FUNDAMENTALLY DIFFERENT from all others.

\textbf{REQUIREMENTS FOR EACH ACTION:}

- Must be contextually appropriate for what you see in the frame

- Must involve subjects/objects actually visible in the frame  

- Must be achievable through image editing (transforming this frame to show the final post-action state)

- Must be UNIQUE -- no two actions should be similar or variations of each other

- Keep actions simple -- avoid complex multi-step actions

\textbf{UNIQUENESS REQUIREMENTS:}

- Each action must involve DIFFERENT types of movement, interaction, or change

- Avoid similar actions like ``wave with right hand'' vs. ``wave with left hand''

- Avoid variations of the same action like ``open door'' vs. ``open door wider''

- Focus on completely different action categories: gestures, movements, interactions, expressions, object manipulations, etc.

\textbf{EXAMPLES OF GOOD DIVERSE ACTIONS:}

- Person waves hand (gesture)

- Person turns head to look left (movement)

- Person picks up object (interaction)

- Door opens (object change)

- Person smiles (expression)

\textbf{EXAMPLES OF BAD SIMILAR ACTIONS (avoid these patterns):}

- Person waves right hand + Person waves left hand

- Person opens door + Person closes door

- Person looks left + Person looks right

\textbf{FOR EACH ACTION, PROVIDE:}

action\_caption: Clear, concise description of what's happening in the action

\textbf{OUTPUT FORMAT (JSON only, no extra text):}

\{

  \hspace{0.3cm}``actions'': [
  
    \hspace{0.6cm}\{``action\_id'': 0, ``action\_caption'': ``...''\},
   
    \hspace{0.6cm}\{``action\_id'': 1, ``action\_caption'': ``...''\},
    
    \hspace{0.6cm}\{``action\_id'': 2, ``action\_caption'': ``...''\},
    
    \hspace{0.6cm}...
  
  \hspace{0.3cm}]
  
\}

Return ONLY the JSON.
 
    \end{tcolorbox}
    \caption{\textbf{Action proposal prompt.} The VLM is instructed to generate $N$ semantically distinct and physically plausible actions conditioned on the anchor frame, which serve as candidate counterfactuals.}
    \label{fig:prompt_a}
\end{figure*}

\begin{figure*}[t]
    \centering
    \begin{tcolorbox}[
        colback=promptbg,
        colframe=promptborder,
        arc=2mm,
        boxrule=1pt,
        width=\linewidth,
        left=10pt, right=10pt, top=10pt, bottom=10pt
    ]
        \ttfamily\small 
You are evaluating proposed actions for a video frame to determine if they are realistic, feasible, and UNIQUE from each other.

\textbf{ORIGINAL VIDEO CONTEXT:} \placeholder{caption}

\textbf{PROPOSED ACTIONS:} \placeholder{actions}

\textbf{EVALUATION CRITERIA:}

1) Subject Presence: Are all subjects/objects mentioned in the action actually visible in the frame?

2) Physical Feasibility: Is the proposed action physically possible given the current frame?

3) Contextual Appropriateness: Does the action make sense in this scene/environment?

4) General Feasibility: Could this action plausibly happen in this scene?

5) UNIQUENESS: Is this action fundamentally different from all other proposed actions?

\textbf{UNIQUENESS EVALUATION:}

- Actions are TOO SIMILAR if they involve the same type of movement/gesture (e.g., ``wave right hand'' vs. ``wave left hand'')

- Actions are TOO SIMILAR if they are variations of the same action (e.g., ``open door'' vs. ``close door'')

- Actions are TOO SIMILAR if they involve the same body part doing similar movements (e.g., ``turn head left'' vs. ``turn head right'')

- Actions PASS uniqueness if they involve completely different types of actions (gesture vs. expression vs. object interaction vs. movement)

\textbf{YOUR TASK:}
For each action, determine if it passes ALL evaluation criteria. An action FAILS if:

- It mentions subjects/objects not visible in the frame

- It's physically impossible given the current pose/position

- It's completely inappropriate for the scene context

- It's unrealistic for this type of scene

- \textbf{It's too similar to another proposed action}

\textbf{RESPONSE FORMAT:}
Provide your response in this exact JSON format:

\{

    \hspace{0.3cm}``evaluations'': [
    
        \hspace{0.6cm}\{
        
            \hspace{0.9cm}``action\_id'': 0,
            
            \hspace{0.9cm}``passed'': True/False,
            
            \hspace{0.9cm}``uniqueness\_check'': ``Explanation of uniqueness evaluation'',
            
            \hspace{0.9cm}``similarity\_issues'': ``List any actions this is too similar to, or None''
            
        \hspace{0.6cm}\},
        
        \hspace{0.6cm}...
        
    \hspace{0.3cm}],
    
    \hspace{0.3cm}``overall\_assessment'': ``Summary of the evaluation results including uniqueness analysis'',
    
    \hspace{0.3cm}``uniqueness\_summary'': ``Overall assessment of how unique the actions are from each other''
    
\}

Analyze the frame and evaluate each proposed action for both realism and uniqueness.
    \end{tcolorbox}
    \caption{\textbf{Action quality evaluation prompt.} This prompt filters proposed actions by enforcing subject presence, physical and contextual feasibility, and semantic uniqueness within the scene.}
    \label{fig:prompt_b}
\end{figure*}

\begin{figure*}[t]
    \centering
    \begin{tcolorbox}[
        colback=promptbg,
        colframe=promptborder,
        arc=2mm,
        boxrule=1pt,
        width=\linewidth,
        left=10pt, right=10pt, top=10pt, bottom=10pt
    ]
        \ttfamily\small 
\textbf{ROLE:}
You are an expert image-editing prompt engineer.

\textbf{GOAL:}
Look at this INITIAL\_FRAME (image) and an ACTION\_CAPTION (text describing what has already happened by the final frame), produce a single, executable edit instruction for an image-editing model that transforms the initial frame into the believable final state, while preserving all unrelated details for video continuity (FFLF).

\textbf{INPUTS:}

- INITIAL\_FRAME: attached the image of the starting frame

- ACTION\_CAPTION: a short description of the event’s outcome by the final frame

\textbf{OUTPUT FORMAT:}

Return ONLY the following JSON:

\{``edit\_prompt'': ``one concise, model-ready edit instruction''\}

\textbf{HOW TO WRITE edit\_prompt:}

1) Start with a precise verb: ADD / REMOVE / REPLACE / MODIFY.

2) Describe the visible final-state evidence of the action (effects on objects, materials, surfaces).

3) Constrain for continuity:

\hspace{0.3cm}- Keep everything else unchanged.
   
\hspace{0.3cm}- Preserve camera, composition, perspective, framing, time of day, grain/noise.

\hspace{0.3cm}- Preserve subject identity (faces, clothing, hairstyle) unless explicitly edited.

\hspace{0.3cm}- Preserve existing text layout (font, size, alignment, perspective) unless explicitly edited.
   
\hspace{0.3cm}- No camera move, no focal length or depth-of-field change.

4) Ensure physical plausibility:

\hspace{0.3cm}- Match lighting direction, shadow softness, reflections, occlusions, contact shadows.

\hspace{0.3cm}- Respect materials (wetness, scorch, dents, debris, splash, smoke) consistent with the scene.

\hspace{0.3cm}- Add subtle motion evidence only if warranted, without altering pose/composition.

5) Pinpoint WHERE the change occurs using anchors from the image (e.g., ``on the desk in front of the mug'', ``upper-left window pane'', ``subject’s right hand'').

6) Include a brief negative clause to prevent artifacts:

\hspace{0.3cm}- ``no extra people/limbs, no duplicates, no distortion, no style drift, no warped text''.

\textbf{STYLE \& BREVITY:}

- 1--3 sentences total.

- Be conservative; prefer minimal edits needed to depict the outcome.

- Do not ask questions; make your best, plausible inference from inputs.

- Output only the JSON object above—no explanations.

\textbf{FEW-SHOT EXAMPLES (follow style, not content):}

\textbf{Example A}

INITIAL\_FRAME: \placeholderb{Man holding a ceramic mug on a wooden desk; window light from left.}

ACTION\_CAPTION: \placeholderb{He spills coffee from the mug onto the desk.}

Expected:
\placeholderb{\{``edit\_prompt'': ``ADD a fresh coffee spill spreading from the mug’s rim onto the wooden desktop with dark wet sheen, small droplets, and a thin run toward the front edge; match left-window lighting for highlights and contact shadows; preserve man, pose, mug, camera, perspective, grain, and desk items. Keep everything else unchanged; no extra hands, no new objects, no distortion.''\}}

\textbf{Example B}

INITIAL\_FRAME: \placeholderb{Storefront banner reading ``SUMMER SALE'', angled perspective.}

ACTION\_CAPTION: \placeholderb{The sale ended; banner now says `WELCOME BACK'.}

Expected:
\placeholderb{\{``edit\_prompt'': ``REPLACE the banner text with `WELCOME BACK', matching the original font, size, color, material, weathering, folds, and perspective warp; keep stitching and fabric texture consistent. Keep everything else unchanged; no warped letters, no background shift, no added objects.''\}}

\textbf{Example C}

INITIAL\_FRAME: \placeholderb{Woman jogging in a park, ponytail; late-afternoon sun from right.}

ACTION\_CAPTION: \placeholderb{She finishes in light rain; she’s slightly wet and breathing.}

Expected:
\placeholderb{\{``edit\_prompt'': ``MODIFY the scene to add a light drizzle and subtle wetness on hair and jacket with fine highlights; add faint condensed breath near the mouth; keep sun direction from right and existing shadowing intact. Keep everything else unchanged; no extra people, no heavy downpour, no camera change.''\}}

\textbf{CALL TEMPLATE (fill and run):}

Use the instructions above.

INITIAL\_FRAME: attached image frame

ACTION\_CAPTION: \placeholder{action\_caption}

Return only:
\{``edit\_prompt'': ``your concise, model-ready edit instruction''\}

    \end{tcolorbox}
    \caption{\textbf{Editing instruction generation prompt.} The high-level action caption is converted into structured image-editing instructions suitable for the image editing model.}
    \label{fig:prompt_c}
\end{figure*}

\begin{figure*}[t]
    \centering
    \begin{tcolorbox}[
        colback=promptbg,
        colframe=promptborder,
        arc=2mm, 
        boxrule=1pt,
        width=\linewidth,
        left=10pt, right=10pt, top=10pt, bottom=10pt
    ]
        \ttfamily\small 
Look at these two images: the original image and the edited image.

\textbf{EDITING INSTRUCTION:} \placeholder{editing\_prompt}

Compare the original and edited images. Has the editing instruction been correctly applied in the edited image?

First, provide your evaluation as either ``YES'' or ``NO''.

- ``YES'' if the editing instruction has been correctly applied.

- ``NO'' if it has not been correctly applied or if the changes are incorrect.

If your evaluation is ``NO'', then provide a short explanation of what is wrong with the generation.

\textbf{OUTPUT FORMAT:}

EVALUATION: YES/NO

EXPLANATION: Only provide if evaluation is NO -- short explanation of what went wrong

    \end{tcolorbox}
    \caption{\textbf{Edited frame evaluation prompt.} The prompt compares the original and edited frames to verify that the intended action is correctly realized while preserving scene identity.}
    \label{fig:prompt_d}
\end{figure*}

\begin{figure*}[t]
    \centering
    \begin{tcolorbox}[
        colback=promptbg,
        colframe=promptborder,
        arc=2mm,
        boxrule=1pt,
        width=\linewidth,
        left=10pt, right=10pt, top=10pt, bottom=10pt
    ]
        \ttfamily\small 
\textbf{ROLE:}
You are an expert image-editing prompt engineer specialized in Qwen-style edit instructions.

\textbf{SITUATION:}
Multiple attempts have failed. Produce ONE new, executable edit instruction that is fundamentally different from all prior attempts.

\textbf{INPUTS:}

- DESIRED OUTCOME: \placeholder{desired\_outcome}

- CURRENT FAILED PROMPT: \placeholder{original\_prompt}

- ALL PREVIOUS FAILED PROMPTS:
\placeholder{failed\_list}\placeholder{failure\_block}

\textbf{PLAYBOOK ADHERENCE (apply silently; do NOT output this section):}

- Choose exactly ONE editing family: Text edit / Local appearance / Global restyle (semantic) / Micro or region edit / Identity control / Poster \& composite / Lighting \& camera.

- Include the continuity clause: ``Keep everything else unchanged''.

- If editing text, preserve font, size, color, alignment, perspective.

- Preserve subject identity (faces, clothing, hairstyle) unless the outcome requires a change.

- Match scene physics: lighting direction, shadow softness, reflections, materials, occlusions, perspective.

- Add a concise negative clause (e.g., ``no duplicates, no distortion, no warped text, no style drift'').

- Pinpoint the location with unambiguous anchors visible in the image.

\textbf{DIFFERENTIATION MANDATE (must differ on $\geq$ 2 axes from ALL failures):}

1) VERB FAMILY: switch to a new one (ADD / REMOVE / REPLACE / MODIFY) not used before.

2) FOCUS: change target (subject $\longleftrightarrow$ environment/background $\longleftrightarrow$ lighting/shadows/reflections $\longleftrightarrow$ layout/spacing).

3) STYLE: change descriptive style (hyper-specific anchors/measurements $\longleftrightarrow$ conceptual brevity; technical terms $\longleftrightarrow$ plain language).

4) SPATIAL ANCHORING: use new landmarks; switch relative $\longleftrightarrow$ absolute positioning.

5) PHRASEOLOGY: avoid reusing distinctive phrases or any 3-word sequence from failed prompts.

\textbf{OUTPUT RULES (hard constraints):}

- Start with EXACTLY one verb: ADD, REMOVE, REPLACE, or MODIFY.

- 1--2 sentences total, $\leq$ 55 words.

- Forbid camera/pose/focal changes unless explicitly required by DESIRED OUTCOME.

- Depict only visible final-state evidence; prefer the minimal change that unambiguously communicates the outcome.

- No meta-commentary, no lists, no JSON, no quotes, no hedging.

RETURN Only the new editing instruction as plain text, in 1--2 sentences ($\leq$ 55 words), starting with the chosen verb.

    \end{tcolorbox}
    \caption{\textbf{Refinement prompt.} When an edit attempt fails, this prompt generates a new image-editing instruction that is explicitly differentiated from prior attempts and targets previously observed failure modes.}
    \label{fig:prompt_e}
\end{figure*}

\end{document}